%% file: main.tex
\pdfoutput=1
\documentclass{article}
\usepackage{microtype}
\usepackage{graphicx}
\usepackage{booktabs}
\usepackage{hyperref}
\usepackage{xcolor}
\usepackage{colortbl}
\definecolor{fsrow}{HTML}{E8F0FB}
\usepackage{amsmath,amssymb,amsthm}
\usepackage{tikz}
\usetikzlibrary{positioning}
\usepackage[preprint]{icml2027}
\hypersetup{pdfauthor={Runguo Li, Dezhen Lu}, pdftitle={FlyingStart: Learned Cross-Token Warm Starts for Fixed-Point Decoding in Recurrent-Depth Language Models}}
\newtheorem{proposition}{Proposition}

\icmltitlerunning{FlyingStart: Learned Warm Starts for Recurrent-Depth Language Models}

\begin{document}

\twocolumn[
\icmltitle{FlyingStart: Learned Cross-Token Warm Starts for\texorpdfstring{\\}{ }Fixed-Point Decoding in Recurrent-Depth Language Models}
\begin{icmlauthorlist}
\icmlauthor{Runguo Li}{uiuc}
\icmlauthor{Dezhen Lu}{tue}
\icmlauthor{Zixiang Ni}{xjtu}
\end{icmlauthorlist}
\icmlaffiliation{uiuc}{University of Illinois Urbana-Champaign}
\icmlaffiliation{tue}{University of T\"ubingen}
\icmlaffiliation{xjtu}{Xi'an Jiaotong University}
\icmlcorrespondingauthor{Runguo Li}{\mbox{runguol2@illinois.edu}}
\vskip 0.3in
]
\printAffiliationsAndNotice{}

\begin{abstract}
Recurrent-depth language models generate each token by iterating a shared block of layers until its hidden state converges.
Inference methods for these models decide \emph{when to stop} the iteration, yet every token still starts from random noise and pays the full cost of converging.
We show that \emph{where the iteration starts} is a second, independent lever.
A token's fixed point is predictable from the state at which the previous token stopped, after a learned change of coordinates, and we introduce \textbf{FlyingStart}, a linear warm start that is fitted in closed form on the frozen model and costs under 5\% of one loop.
The iteration and its stopping rule are unchanged, so the exit criterion certifies the same distance to the fixed point from any start: a better start removes loops at unchanged accuracy.
On Huginn-0125, FlyingStart matches the accuracy of the strongest training-free baseline with 14--19\% fewer decoding layers, raises batched throughput by 16--18\% in two serving engines, and adds a further 10--14\% on top of self-speculative decoding.
A contraction analysis predicts the number of loops saved on all four models we study, including one where the prediction was recorded before any evaluation, and token by token on Huginn, and it identifies when a warm start pays off: the saving grows with the number of loops a model needs to converge.
\end{abstract}

\begin{figure*}[t]
\centering
\includegraphics[width=\textwidth]{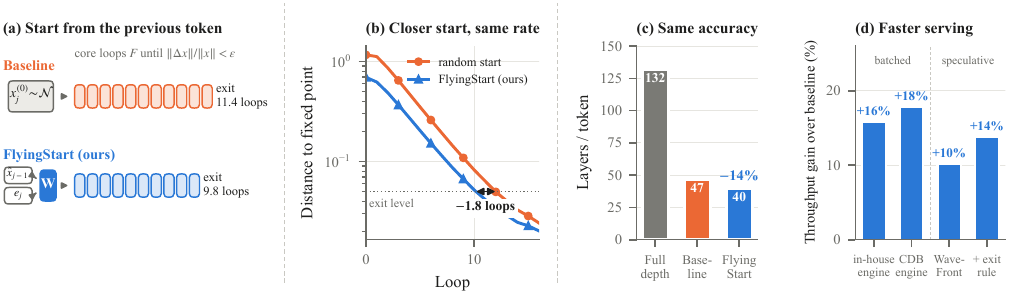}
\caption{\textbf{FlyingStart starts each token closer to its fixed point, so it needs fewer loops for the same accuracy} (Huginn-0125).
(a)~The baseline starts every token from noise; FlyingStart maps the previous token's exit state $\hat x_{j-1}$ and the prelude output $e_j$ through a fitted linear map $W$ to the initial state. Both then run the same iteration and the same exit rule.
(b)~Mean distance to the fixed point over 1,969 generated tokens: both starts contract at the same rate, and FlyingStart reaches the exit level 1.8 loops earlier.
(c)~Layers per token at each method's cheapest setting that keeps full-depth accuracy.
(d)~Throughput gain over the baseline in two batched serving engines, and on top of WaveFront self-speculative decoding at batch size 1.}
\label{fig:teaser}
\end{figure*}

\section{Introduction}

Recurrent-depth (``looped'') language models reuse one block of layers many times per token, trading parameters for test-time compute \citep{geiping2025huginn,ouro2025,mcleish2025retrofit}.
In models trained with a randomized number of iterations, such as Huginn-0125 \citep{geiping2025huginn}, the per-token computation behaves like a fixed-point iteration $x \leftarrow F(x;\,\text{context})$: after enough loops the state stops changing, and the result does not depend on where the iteration started.
The same iteration makes these models expensive to serve. At its full depth of 32 loops, Huginn runs 132 layers for every generated token.

The fixed-point view has shaped how these models are decoded.
Stopping each token once its state has converged, and keeping only the last loop's key--value (KV) entries, is the strongest training-free recipe to date \citep{parallelsamplers2025}; in our reproduction it matches full-depth accuracy at 47 layers per token.
Depth-asynchronous batching \citep{cdb2026} and self-speculative decoding \citep{wavefront2026} then turn the saved loops into throughput.
All of these methods take the iteration as given and decide \emph{when to stop} it or how to schedule it.

This paper asks the complementary question: \emph{where should the iteration start?}
Every token currently starts from random noise and contracts toward its fixed point at a rate set by the model.
On Huginn the error shrinks by a factor of about $0.745$ per loop, so a token needs roughly ten loops whatever the stopping rule.
The contraction rate belongs to the trained weights: solver acceleration, adapters and distillation all leave it unchanged (Section~\ref{sec:analysis}).
The starting point is the lever that inference can still move.

Warm-starting an iteration from a neighbouring solution is a classical idea, from numerical continuation \citep{allgower1990continuation} to deep equilibrium flow estimation, which reuses the previous video frame's fixed point \citep{bai2022deqflow}, and \citet{geiping2025huginn} suggest it for Huginn.
For language the direct version fails: neighbouring tokens have different fixed points, and the previous token's state is no closer to the current fixed point than noise is.
Our key observation is that the relation between neighbouring fixed points is nonetheless simple.
After a \emph{learned linear change of coordinates}, the previous token's state predicts a large part of the next token's fixed point.
\textbf{FlyingStart} turns this observation into a method: one linear map of the previous exit state and the current input, fitted by ridge regression on a few hundred text sequences in minutes, sets the initial state, and the model then iterates and stops exactly as before.
Each token begins its iteration with the progress of the previous one, as a relay runner is already moving at the handover (Figure~\ref{fig:teaser}).

Because FlyingStart changes only the start, the fixed-point view states precisely what to expect from it.
Under a contraction assumption that Huginn satisfies closely, the stopping rule certifies the same error from any start, so an inaccurate prediction costs loops and never accuracy, and a start at distance $d_\text{init}$ instead of $d_\text{rand}$ saves $\log(d_\text{rand}/d_\text{init})/\log(1/\rho)$ loops (Proposition~\ref{prop:warm}).
This law predicts a saving of 1.8 loops per token on Huginn, and we measure 1.7.

\paragraph{Contributions.} This paper makes four contributions.
\begin{itemize}
\setlength{\itemsep}{1pt}\setlength{\parskip}{0pt}\setlength{\topsep}{0pt}
\item \textbf{A new lever for recurrent-depth inference.} FlyingStart is, to our knowledge, the first learned warm start for a released recurrent-depth language model. It is a closed-form linear map on a frozen model, costs under 5\% of one loop, and composes with any stopping rule, batching scheme and speculative decoder (Section~\ref{sec:method}).
\item \textbf{A theory that predicts the gain.} A contraction account shows that the warm start is safe, gives the number of loops it saves, and delimits where it pays off. Its predictions hold on four models, one of them forecast before its evaluation, and token by token (Sections~\ref{sec:theory} and~\ref{sec:analysis}).
\item \textbf{Savings in compute and wall-clock time.} On Huginn, FlyingStart uses 14--19\% fewer decoding layers than the strongest baseline at matched accuracy, or gains up to 3.3 points at an equal budget. It raises throughput by 16--18\% in two serving engines and adds 10--14\% to self-speculative decoding (Section~\ref{sec:results}).
\item \textbf{Evidence that the start is the remaining lever.} Three ways of accelerating the contraction itself fail on Huginn, and starts that are closer on average without being token-specific save no loops (Section~\ref{sec:analysis}).
\end{itemize}

\section{Preliminaries}\label{sec:setup}

\paragraph{Recurrent-depth models.}
A recurrent-depth model maps token $j$ through a \emph{prelude} to an embedding $e_j$, iterates a \emph{core} block $x^{(t+1)}_j = F(x^{(t)}_j, e_j;\, \mathcal{C}_{<j})$ for $t = 0,\dots,T-1$, where $\mathcal{C}_{<j}$ is the KV cache of earlier tokens, and reads out the next-token distribution with a \emph{coda}.
Huginn-0125 (3.5B) has 2 prelude, 4 core and 2 coda layers and is trained with a random number of loops (mean 32), so a token computed with $r$ loops passes through $4 + 4r$ layers.
The retrofitted Llama-3.2-1B and OLMo-2-1B of \citet{mcleish2025retrofit} have 4 prelude, 6 core and 4 coda layers ($8 + 6r$ layers), their TinyLlama-1.1B has 8 core layers ($8 + 8r$), and all are trained with up to 32 loops.
All three draw the initial state $x^{(0)}_j$ at random.

\paragraph{Fixed-point decoding and the strongest baseline.}
Following \citet{parallelsamplers2025}, each token exits when the relative state change $\|x^{(t)} - x^{(t-1)}\| / \|x^{(t)}\|$ drops below a threshold $\epsilon$, and keeps only the KV entries of its last loop.
One KV entry per token is nearly lossless and presents converged keys and values to later tokens, which makes them converge faster (Appendix~\ref{app:extra}).
We call this recipe \emph{the baseline}.
It is the strongest training-free method on Huginn: exits on the KL divergence between consecutive loops \citep{geiping2025huginn} and on output confidence \citep{calm2022} need more loops for the same accuracy (Section~\ref{sec:huginn}).

\section{FlyingStart: choosing where the iteration starts}\label{sec:method}

Autoregressive decoding of a recurrent-depth model solves a sequence of fixed-point problems, one per token, and consecutive problems share almost all of their context.
FlyingStart uses the solution of one problem to start the next.

\subsection{Neighbouring fixed points are linearly related}\label{sec:observe}

\begin{table}[t]
\centering\small
\setlength{\tabcolsep}{3.2pt}
\begin{tabular}{lcccc}
\toprule
& \multicolumn{2}{c}{Huginn} & \multicolumn{2}{c}{Llama} \\
\cmidrule(lr){2-3}\cmidrule(lr){4-5}
Starting point $x^{(0)}_j$ & Wiki & Math & Wiki & Math \\
\midrule
Random (model default) & 1.17 & 1.17 & 1.00 & 1.00 \\
All zeros & 1.00 & 1.00 & 1.00 & 1.00 \\
Copy of previous state $x^\star_{j-1}$ & 1.09 & 1.09 & 1.38 & 1.39 \\
Copy of prelude output $e_j$ & 1.34 & 1.34 & 1.03 & 1.05 \\
\midrule
Linear map of $e_j$ & 0.86 & 0.81 & 0.59 & 0.56 \\
Linear map of $x^\star_{j-1}$ & 0.83 & 0.72 & 0.64 & 0.57 \\
\rowcolor{fsrow}Linear map of both (\textbf{ours}) & \textbf{0.73} & \textbf{0.65} & \textbf{0.52} & \textbf{0.48} \\
\bottomrule
\end{tabular}
\caption{\textbf{The previous token's state predicts the next fixed point only after a learned linear map.} Held-out relative error $\|x^{(0)}_j - x^\star_j\| / \|x^\star_j\|$ of each starting point on Wikipedia and math text (lower is closer).}
\label{tab:fit}
\end{table}

At decoding step $j$, two quantities are available before the first loop: the prelude output $e_j$ and the state $\hat x_{j-1}$ at which the previous token stopped, which lies within about $\epsilon$ of that token's fixed point $x^\star_{j-1}$.
Table~\ref{tab:fit} measures how close each candidate start is to the fixed point $x^\star_j$ on held-out text.

\emph{Raw warm starts fail.}
Copying the previous token's state, the warm start that carries over from continuation methods, is no closer than noise on Huginn (1.09 against 1.17) and farther on the Llama model (1.38).

\emph{A linear map succeeds.}
A linear map of the previous state lowers the error to 0.72--0.83 on Huginn, and adding the prelude output lowers it to 0.65--0.73, and to about 0.5 on the Llama model.
On Huginn this start is as close as the model's own iteration gets after about two loops.

\emph{The signal is token-specific.}
Starting every token at the mean fixed point leaves the error at 0.98--0.99, and a per-dimension affine map of the previous state at 0.91.
The fitted map has an effective rank above one thousand and lies far from the identity (Appendix~\ref{app:details}), so no fixed start, offset or rescaling can stand in for it.

\subsection{The warm-start map}

FlyingStart initializes the iteration of token $j$ with
\begin{equation}\label{eq:fs}
x^{(0)}_j = W z_j, \qquad z_j = [\,e_j;\,\hat x_{j-1};\,1\,].
\end{equation}
The model then iterates from $x^{(0)}_j$ and stops by the same criterion (Algorithm~\ref{alg:fs}).

\begin{algorithm}[t]
\caption{Decoding token $j$ with FlyingStart. The baseline differs only in line~1.}
\label{alg:fs}
\begin{algorithmic}[1]
\REQUIRE prelude output $e_j$, previous exit state $\hat x_{j-1}$, cache $\mathcal{C}_{<j}$, map $W$, threshold $\epsilon$, depth limit $T$
\STATE $x^{(0)} \leftarrow W\,[\,e_j;\,\hat x_{j-1};\,1\,]$ \hfill $\triangleright$ baseline: random $x^{(0)}$
\FOR{$t = 1, \dots, T$}
\STATE $x^{(t)} \leftarrow F(x^{(t-1)}, e_j;\, \mathcal{C}_{<j})$
\STATE \textbf{if} $\|x^{(t)} - x^{(t-1)}\| \le \epsilon\,\|x^{(t)}\|$ \textbf{then break}
\ENDFOR
\STATE $\hat x_j \leftarrow x^{(t)}$; append the last loop's KV entries to $\mathcal{C}$
\STATE \textbf{return} next-token distribution $\mathrm{coda}(\hat x_j)$
\end{algorithmic}
\end{algorithm}

\paragraph{Closed-form calibration.}
$W$ is the ridge-regression solution that maps $z_j$ to the full-depth state $x^\star_j$ on a calibration set,
\begin{equation}\label{eq:ridge}
W = \Big(\sum\nolimits_j x^\star_j z_j^{\top}\Big)\Big(\sum\nolimits_j z_j z_j^{\top} + \lambda I\Big)^{-1},
\end{equation}
with $\lambda$ chosen on held-out sequences.
The calibration set consists of 640 text sequences of 512 tokens, 480 from Wikipedia and 160 from the GSM8K training split; a quarter of it gives the same savings (Appendix~\ref{app:details}).
One pass of the frozen model at full depth accumulates the two sums in Equation~\ref{eq:ridge}; this takes minutes on one GPU and needs no gradient steps.
Calibration sees full-depth states and decoding supplies exit states, and the map's held-out error is the same on both (Appendix~\ref{app:details}).

\paragraph{Cost.}
The map is one matrix--vector product per token, under 5\% of the cost of one loop.
It has $2h^2$ parameters for hidden size $h$ (1.6\% of Huginn), adds no KV memory and leaves the model's weights untouched, so it can be switched on for a deployed model and combined with any exit rule.

\subsection{What a better start buys}\label{sec:theory}

The fixed-point view turns the effect of the start into statements that can be checked.

\begin{proposition}\label{prop:warm}
Fix the context and let the core update $F$ of token $j$ satisfy $\|F(x) - x^\star\| \le \rho\,\|x - x^\star\|$ with $\rho < 1$.
Let $x^{(k)} = F(x^{(k-1)})$ and $d_0 = \|x^{(0)} - x^\star\|$.
\begin{enumerate}
\item[(a)] \emph{Same answer.} $x^{(k)} \to x^\star$ from every start, so the start cannot change what the model converges to.
\item[(b)] \emph{Fewer loops.} $\|x^{(k)} - x^\star\| \le \rho^k d_0$. Lowering $d_0$ from $d_\text{rand}$ to $d_\text{init}$ lowers the number of loops needed to reach any tolerance by $\log(d_\text{rand}/d_\text{init}) / \log(1/\rho)$, up to rounding.
\item[(c)] \emph{Same certificate.} If the exit rule fires at loop $k$, that is $\|x^{(k)} - x^{(k-1)}\| \le \epsilon\,\|x^{(k)}\|$, then $\|x^{(k)} - x^\star\| \le \frac{\rho}{1-\rho}\,\epsilon\,\|x^{(k)}\|$, whatever the start.
\item[(d)] \emph{Adaptive exit needs spread.} If every token starts at the same $d_0$ and contracts at the same exact rate $\rho$, every token reaches a given tolerance at the same loop, so a per-token exit rule reduces to a fixed depth. Its advantage over a fixed depth requires $d_0$ (or $\rho$) to vary across tokens.
\end{enumerate}
\end{proposition}

Parts (a) and (b) are the contraction mapping theorem, (c) follows from $\|x^{(k)} - x^\star\| \le \rho\|x^{(k-1)} - x^\star\| \le \rho(\|x^{(k-1)} - x^{(k)}\| + \|x^{(k)} - x^\star\|)$, and (d) follows from (b) with equality.
The proposition has three consequences for FlyingStart.

\emph{Safety.}
By (a) and (c), the exit rule bounds the distance to the fixed point by the same multiple of $\epsilon$ from the learned start as from noise.
An inaccurate prediction therefore costs loops and never accuracy, and FlyingStart needs no verification pass, in contrast to speculative decoding, which must check every draft.

\emph{A law for the saving.}
By (b), the saving depends on two measurable quantities, the ratio of starting distances and the contraction rate:
\begin{equation}\label{eq:law}
\Delta k = \frac{\log\,(d_\text{rand}/d_\text{init})}{\log\,(1/\rho)}.
\end{equation}
On Huginn, $d_\text{rand} = 1.17$, $d_\text{init} \approx 0.69$ and $\rho = 0.745$ give $\Delta k = 1.8$ loops out of the roughly ten that a token needs.
The saving is a number of loops, so it is a large share of the cost for models that contract slowly and need many loops, and a small share for models that converge in a few.

\emph{Adaptive exit.}
By (d), a per-token exit rule can only beat a fixed depth when tokens differ in how far they start from their fixed points.
A random start places every token at the same distance; a learned start does not.
Section~\ref{sec:analysis} tests each of these predictions.

\section{Experiments}\label{sec:results}

\begin{figure*}[t]
\centering
\includegraphics[width=\textwidth]{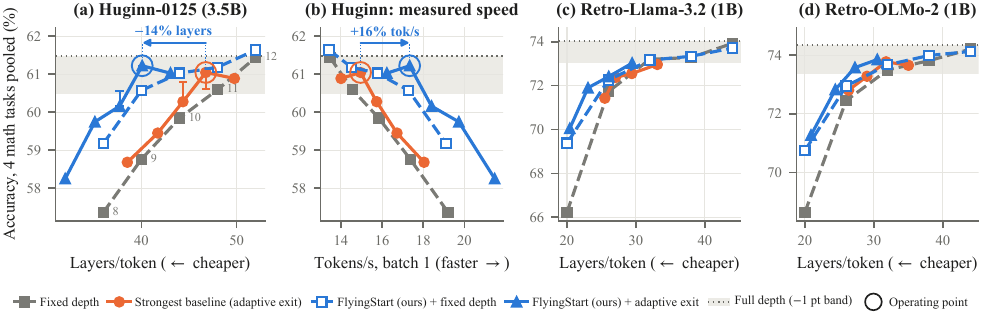}
\caption{\textbf{FlyingStart reaches each accuracy with fewer layers than the strongest baseline on Huginn, and at small budgets on the retrofitted models.}
Accuracy pooled over GSM8K, SVAMP, MultiArith and ASDiv (5,078 problems). The baseline is the state-change exit with one KV entry per token at several thresholds $\epsilon$; FlyingStart applies the same rule from the learned start.
(a)~Huginn; gray numbers are fixed loop counts and vertical bars the range over three seeds.
(b)~The same configurations against measured decoding speed at batch size 1.
(c, d)~The retrofitted models, which reach full-depth accuracy within six loops.
Circles mark each method's operating point, its cheapest configuration not significantly below full depth.}
\label{fig:curve}
\end{figure*}

\subsection{Setup}\label{sec:expsetup}

\paragraph{Models and benchmarks.}
Huginn-0125 is our main model.
The retrofitted Llama-3.2-1B, OLMo-2-1B and TinyLlama-1.1B converge within a few loops and test the boundary that Section~\ref{sec:theory} predicts.
The main benchmarks are chain-of-thought GSM8K (1,319 problems, 8-shot), SVAMP (1,000), MultiArith (600) and ASDiv (2,159); we report accuracy pooled over all 5,078 problems.
Beyond math we evaluate on held-out Wikipedia and code, BBH, IFEval, and zero-shot math and code generation.

\paragraph{Methods.}
We compare FlyingStart with the baseline of Section~\ref{sec:setup} over a grid of thresholds $\epsilon$, and with fixed depths.
All methods run in the same decoding engine with greedy decoding and differ only in the initial state of each token.

\paragraph{Metrics.}
Compute is measured in layers per generated token.
Prefill runs at full depth for every method except in the prefill experiment of Section~\ref{sec:serving}; Appendix~\ref{app:details} gives the end-to-end accounting.
Comparisons are paired over problems, with $z$ values from a sign test and 95\% intervals from a task-stratified bootstrap.
A method's \emph{operating point} is its cheapest configuration whose accuracy is not significantly below full depth, and savings \emph{at matched accuracy} are read off the baseline's Pareto frontier at our accuracy.
We wrote down a pass/fail criterion before each experiment; Table~\ref{tab:prereg} lists all of them with their outcomes.

\subsection{Main results on Huginn}\label{sec:huginn}

\begin{table}[t]
\centering\small
\setlength{\tabcolsep}{3pt}
\begin{tabular}{llcc}
\toprule
Setting & Method & Accuracy (\%) & Layers \\
\midrule
Reference & Full depth (32 loops) & 61.52 & 132.0 \\
\midrule
\multicolumn{4}{l}{\emph{Same threshold: fewer layers}} \\
$\epsilon = 0.07$ & Baseline & 61.07{\scriptsize\,$\pm$\,0.34} & 49.7 \\
\rowcolor{fsrow} & FlyingStart (ours) & 61.34{\scriptsize\,$\pm$\,0.24} & \textbf{43.1} {\scriptsize($-$13\%)} \\
$\epsilon = 0.1$ & Baseline & 60.35{\scriptsize\,$\pm$\,0.13} & 44.3 \\
\rowcolor{fsrow} & FlyingStart (ours) & 60.46{\scriptsize\,$\pm$\,0.23} & \textbf{37.7} {\scriptsize($-$15\%)} \\
\midrule
\multicolumn{4}{l}{\emph{Same budget: higher accuracy}} \\
$\approx$35 layers & Baseline ($\epsilon = 0.2$) & 56.37{\scriptsize\,$\pm$\,0.05} & 34.5 \\
\rowcolor{fsrow} & FlyingStart (ours) & \textbf{59.70}{\scriptsize\,$\pm$\,0.30} & 35.1 \\
10 loops & Fixed depth & 59.85{\scriptsize\,$\pm$\,0.36} & 44.0 \\
\rowcolor{fsrow} & FlyingStart (ours) & \textbf{60.82}{\scriptsize\,$\pm$\,0.25} & 44.0 \\
12 loops & Fixed depth & 61.17{\scriptsize\,$\pm$\,0.22} & 52.0 \\
\bottomrule
\end{tabular}
\caption{\textbf{Main configurations on Huginn.} Accuracy pooled over the four math tasks (mean $\pm$ standard deviation over three seeds) and layers per generated token; in parentheses, the change against the baseline at the same threshold. At $\approx$35 layers FlyingStart runs at $\epsilon = 0.12$.}
\label{tab:main}
\end{table}

\paragraph{Same accuracy, fewer layers.}
FlyingStart moves the whole accuracy--compute frontier toward fewer layers (Figure~\ref{fig:curve}a).
Both methods have their operating point at $\epsilon = 0.085$, where FlyingStart needs 40.1 layers per token and the baseline 46.8, a saving of 14\% that is the same in each of three seeds.
Along the frontier, the saving at matched accuracy is 16.6\% at $\epsilon = 0.1$ (95\% CI $[14.1, 23.5]$) and 18.9\% at $\epsilon = 0.12$ ($[13.0, 20.7]$).
At an equal threshold FlyingStart uses 13--15\% fewer layers at the baseline's accuracy (Table~\ref{tab:main}), as Proposition~\ref{prop:warm}(c) leads us to expect.
The saving also holds when the operating point is selected on held-out problems \citep{haltingselection2026}: over 1,000 random half-splits it is 14.4\% (Appendix~\ref{app:select}).

\paragraph{Same budget, higher accuracy.}
At about 35 layers per token FlyingStart reaches 59.70\%, 3.3 points above the baseline ($z = 7.61$) and 2.7 points above a fixed depth of 8 loops.
At a fixed depth of 10 loops it brings accuracy to within noise of full depth, a level the random start reaches at 12.

\paragraph{Any stopping rule.}
The gain carries over to the KL criterion of \citet{geiping2025huginn}: at equal thresholds FlyingStart saves 1.5--1.6 loops per token with no loss, which amounts to 17--22\% fewer layers at matched accuracy.
Exits on the output distribution \citep{calm2022} need 28--34\% more loops than state change at matched accuracy (Table~\ref{tab:conf}).

\subsection{Generalization}\label{sec:second}

\paragraph{Faster-converging models.}
On the retrofitted Llama and OLMo models, which reach full-depth accuracy within six loops, the curves follow the predicted saving of about 0.7 loops (Section~\ref{sec:analysis}; Figure~\ref{fig:curve}c,d).
At two loops FlyingStart gains 3.2 points on Llama ($z=6.29$) and 2.1 on OLMo ($z=4.31$), and at the smallest budgets it uses 12--15\% and 8--14\% fewer layers at matched accuracy.
From four loops on, the two starts are within noise of each other.
As a forecast test, before evaluating a fourth model, the retrofitted TinyLlama, we recorded that FlyingStart would save 0.71 loops, gain at two loops and match the random start from five on.
It saves 0.56--0.85 loops, gains 3.5 points at two loops ($z = 6.60$) and is within 0.3 points at five and six (Table~\ref{tab:tiny}).
FlyingStart removes the first stretch of the trajectory, which lies inside the operating range on Huginn and matters at tight budgets on fast-converging models.

\begin{figure*}[t]
\centering
\includegraphics[width=\textwidth]{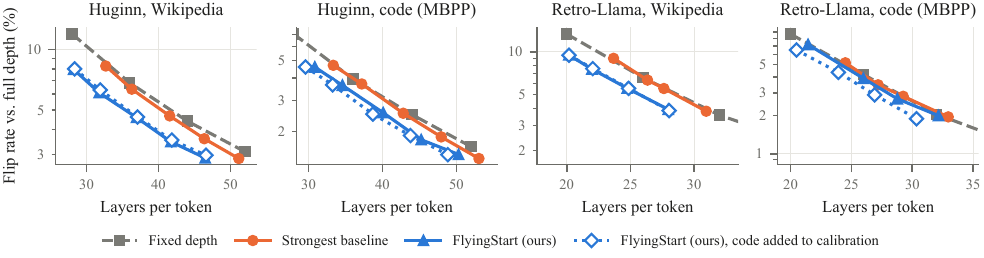}
\caption{\textbf{Outside math, FlyingStart agrees with full depth more often at equal layers.}
Fraction of tokens whose greedy prediction differs from full depth (log scale), under teacher forcing on 128 held-out Wikipedia articles and on MBPP test programs. Hollow diamonds: map calibrated with 160 additional code sequences.}
\label{fig:text}
\end{figure*}

\paragraph{Beyond math.}\label{sec:text}
On held-out Wikipedia and MBPP programs we measure the \emph{flip rate}, the fraction of tokens whose greedy prediction differs from full depth (Figure~\ref{fig:text}).
On Wikipedia, FlyingStart lowers the flip rate at equal layers on both Huginn and Llama, and at a matched flip rate it uses 9--16\% and 5--13\% fewer layers.
On code, which is absent from the calibration set, the gain is smaller, and adding 160 code sequences to the calibration set increases it on both models while leaving Wikipedia unchanged.
On BBH and IFEval, FlyingStart uses 14\% and 25\% fewer layers than the baseline at the operating threshold with accuracy unchanged ($|z| \le 0.29$; Table~\ref{tab:nonmath}).
On zero-shot math and code generation it uses 11--17\% fewer layers at accuracy within noise of the baseline, but a fixed depth of 10 loops already matches full depth there, so the saving holds at an equal threshold and not at matched accuracy (Table~\ref{tab:m3}).

\paragraph{Calibration data.}
Calibration is cheap, and a small amount of in-domain data makes it effective.
With Wikipedia alone, the saving on math at matched accuracy is 7.5\% (95\% CI $[0.3, 13.0]$); with 160 of the 640 sequences drawn from the GSM8K training split it is 17.6\%, on a pool in which three of the four benchmarks contribute no calibration data.

\subsection{Throughput}\label{sec:serving}

\paragraph{Batched serving.}
In our engine at batch size 128, the fastest configuration within noise of full-depth accuracy is a synchronous fixed depth for both methods.
The baseline needs 12 loops and FlyingStart 10, which raises throughput from 997 to 1,155 tokens/s (+16\%; Table~\ref{tab:throughput}).
The open-source CDB engine \citep{cdb2026}, driven by exit trajectories recorded on 1,500 ShareGPT conversations, gives the same picture: 533 against 628 tokens/s (+18\%).
With per-token exit under continuous depth batching, FlyingStart raises CDB's steady-state throughput by 4--5\% at every threshold on this out-of-domain chat workload.

\paragraph{Batch size 1.}
Per-token latency falls from 55.8 to 47.7 ms at fixed depth ($-15\%$), and with adaptive exit the two operating points decode 15.0 and 17.3 tokens/s (+16\%; Figure~\ref{fig:curve}b), so the layer savings translate directly into wall-clock time.

\paragraph{Prefill.}
Prompt positions are processed in parallel, so only the prelude-output part of the map applies there.
This parallel start adds 1.8 points at 8 prefill loops ($z = 3.29$) and is within noise of the random start from 12 loops on (Table~\ref{tab:prefill}), as expected for a start that is not token-specific (Section~\ref{sec:analysis}).
Prefill depth itself is partly redundant: with 24 prefill loops and FlyingStart decoding, accuracy stays within noise of full-depth prefill ($-0.4$ points, $z = -1.33$), and end-to-end layers per uncached 8-shot problem fall by 24\% against the baseline.

\paragraph{Combined with self-speculative decoding.}
WaveFront \citep{wavefront2026} pipelines tokens through the loops: a token at draft depth $d$ proposes the next one, and full-depth verification keeps the output.
We change only the start of each entering token, which is predicted from its left neighbour after that neighbour has run $d$ loops (Table~\ref{tab:spec}).
Drafts from this start are accepted more often, so the fastest draft depth moves down.
FlyingStart adds 10\% at WaveFront's default $d=4$ and 14\% with the state-change exit, at unchanged accuracy ($|z| \le 0.19$), and the gain persists on a second GPU, on Spec-Bench and under sampling (4--24\%; Table~\ref{tab:spec_extra}).

\input{tables/spec}

\subsection{Ablation}\label{sec:ablation}

\begin{figure}[t]
\centering
\includegraphics[width=0.48\textwidth]{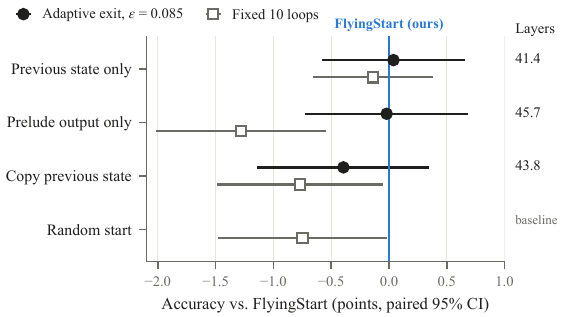}
\caption{\textbf{A learned map of the previous state carries the gain.}
Paired accuracy difference of each variant to FlyingStart in the same setting (Huginn). Right: layers per token at $\epsilon = 0.085$, where FlyingStart needs 40.1 and the random start is the baseline (46.8).}
\label{fig:ablation}
\end{figure}

Figure~\ref{fig:ablation} removes one ingredient of the map at a time.
A map of the previous state alone matches the accuracy of the full map and exits later, at 41.4 instead of 40.1 layers, so we keep both inputs.
The prelude output alone, which corresponds to the input-conditioned initializers of deep equilibrium solvers \citep{bai2022neuralsolvers}, is the weakest learned variant ($-1.28$ points at 10 fixed loops, $z = -3.41$).
Copying the previous state, the warm start suggested by \citet{geiping2025huginn}, is no better than the random start.
More capacity and more history add nothing: an MLP residual trained through the frozen model lands on the same accuracy--layer curve, and the state two tokens back, as a secant predictor would use, changes the held-out error by at most 0.01.
The usable information is the previous token's state after a learned linear change of coordinates, which separates FlyingStart from both the copy warm start and current-input initializers.

\section{Analysis: why the starting point matters}\label{sec:analysis}

\begin{figure}[t]
\centering
\includegraphics[width=0.48\textwidth]{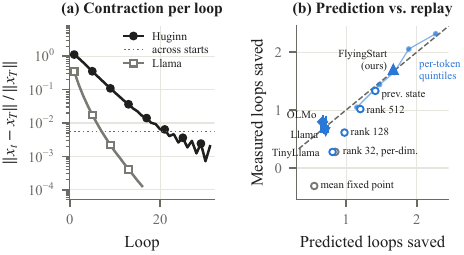}
\caption{\textbf{Huginn contracts at a nearly constant rate, and the contraction law predicts the loops saved by token-specific starts.}
(a)~Relative distance to the final state per loop; dotted: Huginn's spread across random starts.
(b)~Loops saved against the random start, predicted by Equation~\ref{eq:law} and measured on replayed GSM8K tokens. Blue: token-specific maps; gray: starts that are closer on average only.}
\label{fig:conv}
\end{figure}

\begin{figure}[t]
\centering
\includegraphics[width=0.48\textwidth]{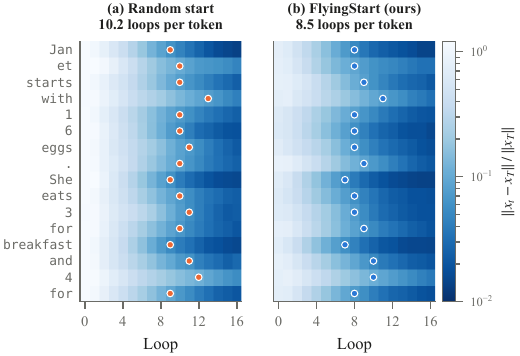}
\caption{\textbf{FlyingStart moves each token's exit earlier and keeps which tokens are hard.} Relative distance to the final state per loop for the first 16 tokens Huginn generates after a GSM8K prompt, from the random start (a) and from FlyingStart (b). Dots mark the exit loop at $\epsilon = 0.085$; Figure~\ref{fig:tokmap} shows a longer span.}
\label{fig:tokmain}
\end{figure}

This section tests the predictions of Section~\ref{sec:theory} against measurements and explains why the start, among the levers we tried, is the one that works.

\paragraph{Huginn is a contraction with a constant rate.}
The per-loop ratio of distances to the fixed point is nearly constant at 0.74--0.75 (Figure~\ref{fig:conv}a), and the fixed point does not depend on the random start.
The measured ratio of the distance at exit to the state change that triggered it is 0.95 from the random start and 0.91 from FlyingStart, well below the bound $\rho/(1-\rho) = 2.9$ of Proposition~\ref{prop:warm}(c): the stopping signal keeps its meaning under FlyingStart.

\paragraph{The law predicts the saving, on average and per token.}
Equation~\ref{eq:law} predicts 1.8 loops on Huginn, and at equal $\epsilon$ we measure 1.7.
The retrofitted models do not contract at a constant rate, so we read the prediction off their random-start trajectories: 0.69 loops on Llama, where we measure 0.6--0.8, 0.67 on OLMo, where we measure 0.8--0.9, and 0.71 on TinyLlama, recorded before its evaluation, where we measure 0.6--0.9.
To test the law per token, we replay 1,969 generated GSM8K tokens on Huginn from each start and read off where each token exits.
Figure~\ref{fig:tokmain} shows one sequence: FlyingStart starts each token at its own distance and moves most exits one or two loops earlier, while the tokens that converge slowly from noise remain the slow ones.
Sorted into quintiles of predicted saving, the measured savings rise from 1.05 to 2.32 loops against a predicted 1.15 to 2.27 (Figure~\ref{fig:conv}b), with a regression slope of 1.16 (95\% CI $[0.97, 1.38]$).

\paragraph{Closer is not enough: the start must be token-specific.}
Starts that are closer than noise on average but carry little token-specific information save fewer loops than the law predicts (gray in Figure~\ref{fig:conv}b).
Starting every token at the mean fixed point costs 0.3 loops \emph{more} than the random start, where the law predicts a saving of 0.55, and an all-zero start also needs more loops for the same accuracy.
A per-dimension affine map or a rank-32 map saves about 0.3 loops instead of the predicted 0.8, and the gap closes as the map gains directions.
We hypothesize that the core contracts at its nominal rate only from states it is used to, namely its training initialization or states near a token's own fixed point.
Of the starts we tested, the learned token-specific map is the one that is both closer and typical.

\paragraph{The contraction rate is fixed after training.}
We tried three ways to speed up the contraction itself on Huginn.
Anderson acceleration \citep{anderson1965} and simpler extrapolations were worse than the plain iteration or diverged: consecutive increments point in opposing directions for the first four loops, so there is no slow subspace to extrapolate along.
A LoRA adapter trained to jump several loops improves the first step only, and every later step still contracts at 0.70--0.75; a few-step student obtained by full fine-tuning loses 4--5 points at the same depth.
For a released model the contraction rate is fixed, and the starting point is the degree of freedom that remains.

\paragraph{Adaptive exit needs an informative start.}
From a random start, per-token adaptive exit is no better than a fixed depth with the same mean number of loops (within 0.04 points).
Proposition~\ref{prop:warm}(d) explains why: the initial relative distance from a random start is $1.166 \pm 0.008$ across tokens, so every token starts equally far from its fixed point.
From the learned start it is $0.699 \pm 0.087$, and adaptive exit then beats a fixed depth at equal mean layers ($+0.87$ points at 40 layers, $z=2.39$).
An informative start therefore also restores the value of the stopping rule.

\section{Related work}

\paragraph{Early exit and serving for looped models.}
\citet{geiping2025huginn} exit on the KL divergence between consecutive loops; \citet{parallelsamplers2025} introduce the state-change criterion with a size-one cache (our baseline) and a diffusion-forcing sampler that trades parallel positions for latency (Appendix~\ref{app:ds}).
Continuous Depth Batching \citep{cdb2026} and Ragged KV \citep{raggedkv2026} make per-token exit efficient in batched serving, and LoopDuet \citep{loopduet2026} uses separate fixed budgets for prefill and decode.
Late Entry \citep{lateentry2026} fine-tunes a model so that a new token runs only its last loops over a context completed to full depth; without fine-tuning, FlyingStart adds 2.0 points at 8 loops in that setting (Appendix~\ref{app:late}).
Under batching, a halting head adds little \citep{haltingheads2026}, in line with our finding that synchronous fixed depth is fastest.
All of these decide when to stop or how to schedule, as do adaptive computation in recurrent networks \citep{graves2016act,banino2021pondernet} and early exit in standard transformers \citep{calm2022,elhoushi2024layerskip}. None changes where the iteration starts, and each combines with FlyingStart.

\paragraph{KV compression and speculative decoding.}
FlashLoop \citep{flashloop2026}, ResidualQuant \citep{residualquant2026}, LoopKV \citep{loopkv2026} and LoopVQ \citep{loopvq2026} compress per-loop KV caches.
BeNice! \citep{benice2026} trains models to share the final loop's cache across depth, which Huginn allows without training.
Self-speculative methods \citep{loopspec2026,wavefront2026,loopfill2026,loopskim2026,loopcache2026} draft with shallow loops and verify at full depth, cutting batch-one latency at the price of more loop evaluations per token.
Neither line reduces the number of loops a token needs, and FlyingStart adds to WaveFront's speedup (Section~\ref{sec:serving}).

\paragraph{Models that change training.}
Attractor Models \citep{attractor2026}, Fixed-Point Reasoners \citep{fpr2026}, Looped Models Done Right II \citep{loopedright2026} and LSTM-style looped transformers \citep{lstmloop2026} build solver-style iteration, shared caches or depth priors into training and require pretraining from scratch.
\citet{loopedright2026} also distill a student that predicts the final state during prefill, at a loss in accuracy; FlyingStart predicts only the start and leaves the iteration unchanged.
Parallel Loop Transformers \citep{plt2025,howtoloop2026,infuseloop2026} are trained so that a token's next loop reads the previous token's hidden state; FlyingStart passes the previous token's state into a frozen model, as the start of its unchanged iteration.
In controlled pretraining, \citet{loopedright1_2026} find that a random initial state gives mixed or negative returns; we keep the trained model and change only the start at inference.

\paragraph{Warm starts, fixed-point solvers and feature prediction.}
Continuation methods solve a sequence of related fixed-point problems by starting each from a prediction based on the previous solutions \citep{allgower1990continuation}; autoregressive decoding of a looped model is such a sequence, indexed by token position.
Learned warm starts for fixed-point optimization predict the starting point from the problem's parameters with a neural network \citep{sambharya2024l2ws}; ours is a closed-form linear map whose main input is the neighbouring problem's solution.
Learned initializers and accelerated solvers are standard for deep equilibrium models \citep{bai2019deq,bai2022neuralsolvers}; they condition on the current input, which is our weakest variant, and standard acceleration fails on the models we study.
Jacobi decoding \citep{song2021parallel,santilli2023jacobi} casts decoding as a fixed-point problem across positions of a standard model.
Speculative decoding methods such as EAGLE \citep{li2024eagle} predict the next token's features to draft tokens; we use a predicted state as the starting point of the exact iteration, so no verification is needed.
Looping ordinary language models without training fails because they do not contract \citep{trainingfreeloop2026}.

\section{Conclusion}
Inference methods for recurrent-depth models choose when to stop iterating; FlyingStart shows that where to start is a separate lever, and a cheap one.
A linear map of the previous token's state and the current input, calibrated in closed form within minutes, removes one to two loops per token on a frozen model, keeps the exit rule's certificate, and adds to stopping rules, batching and self-speculative decoding.
A contraction law predicts the saving from two measurable quantities and tells practitioners in advance which models benefit most: those that need many loops to converge.

\paragraph{Limitations.}
FlyingStart accelerates decoding; in prefill, where positions are processed in parallel, only its input-conditioned part applies, and it helps only when prefill is shallow.
It applies to models whose iteration converges to a start-independent fixed point, such as Huginn and the retrofitted models of \citet{mcleish2025retrofit}; looped models that do not converge, such as Ouro \citep{ouro2025}, have no such fixed point to predict.
The map is calibrated per model on some in-domain data, and on zero-shot math and code generation its saving holds at an equal threshold only.
Token-specific starts for parallel prefill, and training models whose fixed points are easier to predict, are natural next steps.\label{bodyend}

\section*{AI Use Statement}
This manuscript was drafted by the authors. We used Claude Code for two purposes: (i) language editing of the manuscript text and (ii) assistance in implementing and debugging the measurement scripts. It was not used to generate figures, experimental results, related-work content, or research ideas. Its use was limited to sentence-level phrasing and code refactoring. All scientific content, system design, experiments, and analysis were developed by the authors, who verified all modified code and text.

\section*{Impact Statement}
FlyingStart lowers the computation needed to decode from recurrent-depth language models, and with it the energy and hardware cost of serving them.
It changes neither the model's weights nor its stopping rule, and on the tasks we test its accuracy stays within noise of the unmodified model, so it adds no risks beyond those of the underlying models.
Its savings depend on the data used to fit the map and are smaller on formats outside that data, as we report for zero-shot and code generation.

\bibliographystyle{icml2027}
\bibliography{refs}

\appendix
\section{Experimental details and full statistics}\label{app:details}

This appendix gives the implementation details and the complete statistics behind the summary numbers of the main text, in the order of its sections.

\paragraph{Decoding engine.}
All methods run in the same PyTorch decoding engine (bfloat16, greedy decoding), which stores one KV entry per token per layer and reproduces the official forward pass (argmax agreement 100\% on 384 positions in float32).
The baseline and FlyingStart differ only in the initial state.

\paragraph{Decoding compute and end-to-end cost.}
Compute is measured as layers per generated token, which counts decoding only; all methods prefill at full depth except in Table~\ref{tab:prefill}.
On the few-shot math benchmarks prefill dominates FLOPs (795 prompt against 67 generated tokens per problem), and 741 of the prompt tokens are a shared prefix that serving systems compute once.
With the prefix cached, decoding is about a third of the FLOPs and FlyingStart saves about 6\% end to end; with uncached few-shot prompts the end-to-end saving is below 1\%.
At batch size 1 decoding is roughly 95\% of the request time.

\input{tables/prefill}

\paragraph{Prefill depth and start.}
Table~\ref{tab:prefill} varies the number of prefill loops $T_p$ with FlyingStart decoding at $\epsilon = 0.085$; decoding needs 38.0--40.1 layers per token at every $T_p$, so a shallower prompt does not slow down decoding.
In prefill, all positions of a block run the same loop, and Huginn contracts more slowly than in decoding: on held-out GSM8K text the random-start prompt state is still 0.24 from its 32-loop value after 16 loops and 0.12 after 24, a per-loop ratio of about 0.91 against 0.745 in decoding.
The parallel start (relative distance 0.82 on math, against 1.17 for the random start) shifts the mean prompt trajectory by 1.3--2.1 loops, while Equation~\ref{eq:law} predicts 3.8; at the task level its gain at 8 prefill loops corresponds to about 1.2 loops of the random start's accuracy slope.
A start that includes the left neighbour's prelude output has nearly the same held-out error (0.817 against 0.834), so we did not run it.

\paragraph{Hardware.}
Accuracy on the same configuration differs by up to 0.6 points between GPU types, so every comparison is made within one GPU type.
The accuracy--compute curves of Figure~\ref{fig:curve}a, the operating points and Figure~\ref{fig:teaser}c use seed 0 on RTX 3090; Table~\ref{tab:main}, Table~\ref{tab:seeds} and the ablation use H100.
Batched throughput is measured on an H100 in our engine and on an RTX 6000 Ada in the CDB engine; batch-one speed and WaveFront on an RTX 3090.
BBH and TinyLlama run on RTX A4500 and IFEval on RTX 3090.

\paragraph{Calibration of the map.}
The map is fitted on 480 Wikipedia and 160 GSM8K-train sequences of 512 tokens (327k tokens).
Fitting uses full-depth states as inputs while decoding supplies exit states; for the previous-state part of the map the held-out error is the same on both (0.831 vs.\ 0.831 on Wikipedia, 0.723 vs.\ 0.722 on math), and refitting on exit states gives the same error.
The map costs one $2h \times h$ matrix product per token, 3.6\% of one loop on Huginn and 4.6\% on the Llama model, and has $2h^2$ parameters (56M on Huginn, 1.6\% of the model).
For comparison with Table~\ref{tab:fit}, iterating Huginn from a random start for 3 loops gives a relative error of 0.59, and the Llama model after one loop gives 0.35; the linear map is therefore worth about two loops on Huginn and under one loop on the Llama model.

\paragraph{Structure of the map.}
On Huginn, starting every token at the mean fixed point gives a relative error of 0.99/0.98 on Wikipedia/math, a per-dimension affine map of the previous state 0.91, a full map of the previous state 0.83/0.72, and adding the prelude output 0.73/0.65; keeping only the top 128 or 512 directions of the fitted values gives 0.92/0.82 and 0.87/0.77.
The map is far from low rank: the participation ratio $(\sum\sigma_i^2)^2/\sum\sigma_i^4$ of its previous-state block is 1,180 of 5,280 dimensions, the top 512 singular directions hold half of its energy, and it is far from the identity (relative distance 0.85).
Which tokens are predicted well depends on the domain.
On held-out math, digits and punctuation start closer (0.60) than word-initial tokens (0.72), and tokens the model finds predictable start much closer: by quintile of full-depth surprisal the error rises from 0.57 to 0.72, which Equation~\ref{eq:law} converts into 2.5 versus 1.8 loops saved.
On Wikipedia the error is nearly uniform across token types and surprisal (0.71--0.76).

\paragraph{Huginn, main results (Section~\ref{sec:huginn}).}
Figure~\ref{fig:curve}a shows both methods on the same grid: state-change exit at $\epsilon \in \{0.07, 0.085, 0.1, 0.12, 0.15\}$ and fixed depths 8--12.
The cheapest configuration of each method that is not significantly below full depth is $\epsilon = 0.085$ for both: 40.1 layers per token for FlyingStart ($-0.26$ points, 95\% CI $[-1.00, +0.48]$) and 46.8 for the baseline ($-0.43$, $[-1.19, +0.33]$); the next cheaper setting of each method is significantly below full depth ($z=-2.20$ and $-3.02$).
With two more seeds the pair keeps the same layer counts (14.2--14.3\% fewer at every seed), and over the three seeds FlyingStart is 0.29 points more accurate ($z=+0.92$); the baseline at $\epsilon = 0.085$ then falls significantly below full depth ($-0.85$ points, $z=-2.66$) and FlyingStart does not ($-0.56$, $z=-1.51$).
Inverting the baseline's Pareto frontier at our accuracy gives 16.6\% fewer layers at $\epsilon=0.1$ (95\% CI $[14.1, 23.5]$) and 18.9\% at $\epsilon=0.12$ ($[13.0, 20.7]$) over three seeds, and 8.1\% with fixed depth; on the grid of Figure~\ref{fig:curve}a the same inversion gives 14.8\% and 19.1\%.
At about 35 layers FlyingStart reaches 59.70\%, against 56.37\% for the baseline ($z = 7.61$) and 57.00\% for a fixed depth of 8.
At equal $\epsilon$ it uses 1.7 fewer loops per token at the same accuracy ($+0.18$ to $+0.27$ points, $z \le 0.73$).
In Figure~\ref{fig:curve}b, the 95\% bootstrap intervals of the batch-one speeds (100 problems) are narrower than the markers, and the speed ratio of the two operating points is 1.16 $[1.15, 1.17]$; full depth decodes 5.4 tokens/s.

\paragraph{Other stopping rules.}
With the KL criterion of \citet{geiping2025huginn} (thresholds $5\cdot10^{-4}$ to $5\cdot10^{-3}$), FlyingStart uses 1.5--1.6 fewer loops at equal threshold with no loss (all $|z| \le 1.05$) and 16.7--21.7\% fewer layers at matched accuracy (95\% CIs within $[13.0, 26.0]$); this criterion is weaker than state change on Huginn (58.4\% at 9.6 loops).
Exits on the output distribution are weaker still (Table~\ref{tab:conf}): Huginn's first loops give confident wrong predictions, and even with a minimum of 8 loops these exits need 28--34\% more loops than state change at matched accuracy.

\paragraph{Faster-converging models (Section~\ref{sec:second}).}
In float32 the Llama model's distance to the fixed point falls to 0.033 after 4 loops and 0.004 after 8, and the fixed point does not depend on the random start.
FlyingStart gains $+3.15$ points at two loops ($z=6.29$) and $+0.53$ at three, $+1.20$ points against the baseline frontier on average (95\% CI $[+0.89, +1.41]$), with 12--15\% fewer layers at matched accuracy at the two smallest budgets.
From four loops on there is no difference ($-0.04$, $+0.06$, $-0.22$ at 4, 5, 6 loops), and six loops already reach full-depth accuracy.
On OLMo, fitted and evaluated with the same settings, FlyingStart gains $+2.11$ points at two loops ($z=4.31$) and $+0.49$ at three ($z=1.35$), lies $+0.77$ points above the baseline frontier on average (95\% CI $[+0.50, +0.96]$; $+1.09$, $[+0.71, +1.43]$, over its four adaptive points) and uses 8--14\% fewer layers at matched accuracy at the three smallest budgets (all CIs above zero).
From four loops on the differences are within $\pm 0.2$ points, and six loops are within 0.14 points of full depth.
The grids are $\epsilon \in \{0.1, 0.15, 0.2, 0.3\}$ and fixed depths 2--6; full depth is 200 layers.

\paragraph{TinyLlama, a forecast (Section~\ref{sec:second}).}
For the retrofitted TinyLlama-1.1B (4 prelude, 8 core and 4 coda layers; 264 layers at full depth) we first checked our engine against the official forward pass (argmax agreement 100\% on 384 positions in float32; the fixed point does not depend on the random start, relative difference below $10^{-4}$), fitted the map with the default calibration set and measured its starting distance on 2,048 tokens generated after 32 GSM8K prompts (0.54, against 1.00 for the random start).
The random-start trajectory reaches 0.54 after 0.71 loops, which we recorded as the predicted saving together with the criterion that the saving at equal threshold lies within $0.71 \pm 0.3$ loops at each $\epsilon$, a gain of at least one point at two loops, and no difference at five and six loops.
All three held (Table~\ref{tab:tiny}): the saving is 0.56, 0.59, 0.59 and 0.85 loops at $\epsilon = 0.1$, 0.15, 0.2 and 0.3; FlyingStart gains $+3.49$ points at two loops ($z = 6.60$) and $+0.73$ at three ($z=1.80$), and differs by $-0.12$, $-0.22$ and $-0.26$ points at four to six loops (all $|z| \le 1.13$).
Over its six adaptive and small fixed-depth configurations it lies $+1.89$ points above the baseline frontier (95\% CI $[+1.52, +2.25]$) and uses 12--19\% fewer layers at matched accuracy.

\paragraph{Held-out text and code (Section~\ref{sec:second}).}
Sequences have a 128-token context and 256 decoded tokens.
On Wikipedia FlyingStart lowers the flip rate by 1.73 points on Huginn and 1.86 on Llama against the baseline frontier (bootstrap intervals above zero), with Huginn at 12 fixed loops the only point where the two are level (within 0.04); at matched flip rate the adaptive configurations use 9--16\% fewer layers on Huginn and 5--13\% on Llama.
On code the gain is 0.61 and 0.21 points, and adding 160 code sequences to the fit raises it to 0.86 on both models while Wikipedia changes by at most 0.07.
Because teacher forcing hides error accumulation, we also let each configuration generate 255 tokens greedily on Huginn and count the tokens produced before first differing from full depth.
At equal loops FlyingStart agrees longer, by 14.7 tokens on Wikipedia (95\% CI $[7.6, 20.8]$; baseline frontier 16--62 tokens) and 17.6 on code ($[6.2, 27.5]$); at $\epsilon=0.07$, 49.0 Wikipedia tokens against 28.4 for the baseline.

\paragraph{Zero-shot and code generation.}
On zero-shot GSM8K and MATH-500, where decoding dominates, and on MBPP and HumanEval (2,482 problems; Table~\ref{tab:m3}), a fixed depth of 10 loops already matches full depth on Huginn ($+0.20$ points).
FlyingStart at $\epsilon=0.085$ uses 11\% fewer layers than that point ($-0.73$ points, $z=-1.12$) and at equal $\epsilon=0.1$ 17\% fewer than the baseline ($-0.52$, $z=-0.75$); at 10 fixed loops it is 1.21 points below the baseline ($z=-1.85$), so these tasks show a saving at equal threshold and none at matched accuracy.
None of these formats is in the fit, and adding 500 sequences in them, chosen after seeing these results, recovers no gain ($-1.37$ points at 10 fixed loops, $z=-1.98$; $-1.73$ at $\epsilon=0.085$, $z=-2.61$), so the format of the fitting data does not explain it.

\paragraph{BBH and IFEval.}
On BBH (27 tasks, 3-shot chain of thought, 6,511 problems) and IFEval (541 zero-shot prompts), FlyingStart at $\epsilon=0.085$ uses 14\% and 25\% fewer layers than the baseline at the same threshold ($-0.11$ and $+0.18$ points, $|z| \le 0.29$; Table~\ref{tab:nonmath}), which passes the criterion fixed before the run; at $\epsilon=0.1$ on IFEval it is 2.8 points below the baseline ($z=-2.54$).
Every configuration from 10 loops on is within noise of full depth, so on these benchmarks we report savings at equal threshold.

\paragraph{Calibration data.}
Refitting without the GSM8K training problems (Wikipedia only, same number of sequences) raises the map's held-out error on math from 0.65 to 0.87; on the four-task pool it then saves 7.5\% of layers at matched accuracy at $\epsilon = 0.085$ (95\% CI $[0.3, 13.0]$), against 17.6\% for the default fit on the same hardware, and nothing at 10 fixed loops.

\begin{figure}[t]
\centering
\includegraphics[width=0.48\textwidth]{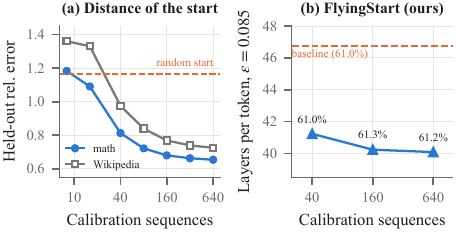}
\caption{\textbf{A quarter of the calibration data gives the full saving} (Huginn). (a)~Held-out relative error of the start against the number of calibration sequences (512 tokens each); dashed: random start. (b)~Layers per generated token at $\epsilon = 0.085$ on the four math tasks, with pooled accuracy; dashed: the baseline at the same threshold.}
\label{fig:calsize}
\end{figure}

\paragraph{Amount of calibration data.}
Figure~\ref{fig:calsize} refits the map on the first $n$ of the 640 calibration sequences after a random shuffle, so about a quarter of each subset comes from GSM8K, with $\lambda$ chosen as before.
The held-out error drops below the random start's from 40 sequences on and changes by less than 0.03 beyond 160.
At $\epsilon = 0.085$ the map from 160 sequences needs 40.2 layers per token at 61.28\% (full map: 40.1 layers, 61.22\%; baseline: 46.8 layers, 61.05\%; $z = 0.63$ against the baseline), and the map from 40 sequences, about 20k tokens, still needs 12\% fewer layers than the baseline at the same accuracy (61.03\%).
Equation~\ref{eq:law} applied to the held-out errors predicts savings of 1.22, 1.83 and 1.96 loops for 40, 160 and 640 sequences; we measure 1.38, 1.63 and 1.67.
Adding 160 ShareGPT conversations to the calibration set leaves the math results unchanged (61.40\%, 40.3 layers).

\paragraph{Throughput (Section~\ref{sec:serving}).}
In our engine at batch size 128 (Table~\ref{tab:throughput}), the baseline at 12 fixed loops reaches 997 tokens/s and FlyingStart at 10 loops 1,155 tokens/s (60.82\% vs.\ 61.17\%, $z=-0.53$); one loop fewer is significantly below full depth for both ($z=-2.39$ and $-2.20$).
Adaptive exit saves more layers, and turning uneven depths into throughput requires depth-asynchronous batching \citep{cdb2026}; in our engine it runs slower than synchronous fixed depth (871 tokens/s for FlyingStart at $\epsilon=0.085$, 732 for the baseline at $\epsilon=0.07$).
At batch size 1 the per-token latency is 47.7 ms with FlyingStart at 10 loops against 55.8 ms at 12 (full depth 138 ms).
The CDB engine \citep{cdb2026} is unchanged except that the map's matrix product is added to every decode step; we replay per-token exit trajectories recorded on 1,500 ShareGPT conversations at width 64.
Fixed depth with continuous batching is again the fastest: 533 tokens/s at 12 loops for the baseline and 628 at 10 loops with FlyingStart.
With adaptive exit at equal $\epsilon \in \{0.07, 0.085, 0.1, 0.15\}$, FlyingStart raises steady-state throughput by 4.0--5.0\% over the random start (e.g.\ 428 vs.\ 409 tokens/s at $\epsilon = 0.1$, mean depth 9.55 vs.\ 10.44 loops) and by 6\% over a zero start.
The adaptive-exit gain follows the loops saved, 0.9 per token on this out-of-domain chat data (1.7 on GSM8K), while prelude, coda and scheduling cost stay fixed.
Adding 160 ShareGPT conversations, disjoint from the replayed requests, to the calibration set raises the saving on this workload to 1.14--1.18 loops per token (0.88--0.90 with the default map); replayed in the same job, the gain over the random start grows from 4.3--5.0\% to 6.2--7.0\%.
We had required at least 2 points more than the default map at three of the four thresholds and measured 1.7--2.0, reaching 2 only at $\epsilon = 0.15$: calibrating on serving traffic helps, but by less than that bar.
For WaveFront, draft depths are chosen on training problems; an H100 gives gains of 12\% and 18\%, out-of-domain Spec-Bench 4\% and 24\% (speed only), and speculative sampling at temperature 0.7 13\% (Table~\ref{tab:spec_extra}).

\paragraph{Ablation (Section~\ref{sec:ablation}).}
A map of the previous state alone differs from the full map by $+0.04$ and $-0.14$ points in the two settings of Figure~\ref{fig:ablation}, and in the token-level replay of Section~\ref{sec:analysis} it saves 1.30--1.33 instead of 1.66--1.70 loops.
A linear map plus an MLP residual trained through the frozen model lands on the same accuracy--layer curve on GSM8K (paired $z = -1.3$, $-1.4$, $+0.8$ at equal $\epsilon$).

\paragraph{Analysis (Section~\ref{sec:analysis}).}
Figure~\ref{fig:conv}a uses 64 generated tokens after each of 8 GSM8K prompts in float32, with final state at $T=32$; Figure~\ref{fig:conv}b uses $\epsilon = 0.07$ and $\rho = 0.733$ from the replay on Huginn, and the range over four $\epsilon$ on Llama, OLMo and TinyLlama.
On average over the 1,969 replayed tokens FlyingStart saves 1.66--1.70 loops against 1.67 predicted; the quintiles of predicted saving are 1.15, 1.47, 1.68, 1.89, 2.27 against 1.05, 1.44, 1.65, 2.05, 2.32 measured (bootstrap over prompts), and the per-token correlation is 0.28--0.33 because loop counts are integers.
On Llama the start at $d_\text{init} \approx 0.48$ is reached by the random-start trajectory after 0.69 loops, and on OLMo the start at 0.54 after 0.67 loops; on TinyLlama the start at 0.54 is reached after 0.71 loops, a value we recorded before its evaluation.
An all-zero start ($d_0 = 1.00$; predicted saving 0.5 loops) uses 0.26 more loops at equal $\epsilon$ for the same accuracy on the four-task pool ($-0.04$, $-0.47$, $-0.21$ points at $\epsilon = 0.1$, $0.07$, $0.15$; all $|z| < 1.6$) and is 1.5--2\% slower in the CDB engine.
A per-dimension affine map or a rank-32 map saves 0.25--0.29 loops instead of the predicted 0.81--0.85, and at rank 512 the map saves 1.0 against 1.2 predicted; on the Llama model the fixed starts are again slower than random ($-0.6$ loops).
Anderson acceleration was run with several window sizes, damping and delayed starts, and the two simpler extrapolations go along the last increment and by over-relaxation; leaving the LoRA adapter on moves the fixed point, and the few-step student (6--8 loops) stops contracting about 0.2 away from the fixed point.
Without a warm start, adaptive exit differs from a fixed depth with the same mean by $+0.04$ points ($\epsilon=0.07$ vs.\ 12 loops) and $-0.02$ ($\epsilon=0.1$ vs.\ 10 loops).
Over 977 replayed GSM8K tokens the initial relative distance from a random start has 10th--90th percentiles 1.156--1.176, and from the learned start 0.595--0.806; with the learned start adaptive exit beats fixed depth by $+0.87$ points at 40 layers ($z=2.39$) and $+0.61$ at 43--44 layers ($z=1.87$).

\section{Additional results}\label{app:extra}

\input{tables/huginn_tasks}
\input{tables/retro_tasks}
\input{tables/tiny_tasks}
\input{tables/text}
\input{tables/throughput}
\input{tables/m3}
\input{tables/nonmath}
\input{tables/spec_extra}

\paragraph{Relation to late entry.}\label{app:late}
Late Entry \citep{lateentry2026} lets a new token run only its last $K$ loops over a context completed to full depth, which it obtains by fine-tuning and a pipelined cache; its released Huginn checkpoint was not available to us, so we reproduce the mechanism without training (Table~\ref{tab:late}).
Every token emits after $K$ loops; the settings differ only in the context later tokens read.
With per-loop KV truncated at $K$, the early-exit baseline of \citet{lateentry2026}, accuracy at 8 loops is 34.9\%; keeping each token's state after $K$ loops as its single KV entry, our baseline, gives 57.4\% at the same cost, and completing every token to 32 loops adds only 0.85 points more ($z = 1.88$) at 132 instead of 36 layers per token.
FlyingStart adds 1.8 points on the depth-$K$ context ($z = 3.8$) and 2.0 on the completed one ($z = 4.6$; criterion fixed before the run), so at 8 loops it gains more than completing the context does, and the two combine; from 12 loops on all deep-context settings are within noise of full depth.

\input{tables/late}

\paragraph{Exits on the output distribution.}\label{app:conf}
We test the usual confidence signals of early-exit transformers \citep{calm2022} as stopping rules on Huginn, with the random start and one KV entry per token as in our baseline: the maximum probability, the entropy and the gap between the two largest probabilities of the next-token distribution, checked after every loop from a minimum number of loops on (Table~\ref{tab:conf}).
Thresholds and minimum depths were chosen on 100 GSM8K training problems by a rule fixed before the test run; the minimum depth was added after a first pilot in which, without it, many tokens stopped after 2--4 loops with confident wrong predictions.
Without a minimum or with a minimum of 4, every signal is less accurate than a fixed depth of 8 loops.
With a minimum of 8, the three signals reach 59--60\% and need 28--34\% more loops than the state-change frontier at the same accuracy, or 45--56\% more layers when the coda pass for each check is counted; state change therefore remains the strongest training-free baseline.

\input{tables/conf}

\paragraph{One KV entry per token.}
On Huginn, decoding at full depth while keeping only the last loop's KV entries gives 43.4\% on the first 500 GSM8K problems, 1.0 point below keeping all 32 loops ($z = 1.0$); keeping 2, 4 or 8 loops makes no difference in token disagreement (0.70\%, 0.65\%, 0.65\% vs.\ 0.64\%).
This reproduces \citet{parallelsamplers2025}.

\paragraph{Converged context speeds up convergence.}
With KV entries aligned by loop (each loop reads the same loop of earlier tokens), a new token needs on average 17.9 loops to reach the KL criterion of \citet{geiping2025huginn}; reading the last loop of earlier tokens it needs 11.1 (GSM8K; 18.1 vs.\ 12.7 on Wikipedia).
The relative distance to the final state after 10 loops is 0.22 with aligned and 0.074 with last-loop context.

\begin{figure*}[t]
\centering
\includegraphics[width=\textwidth]{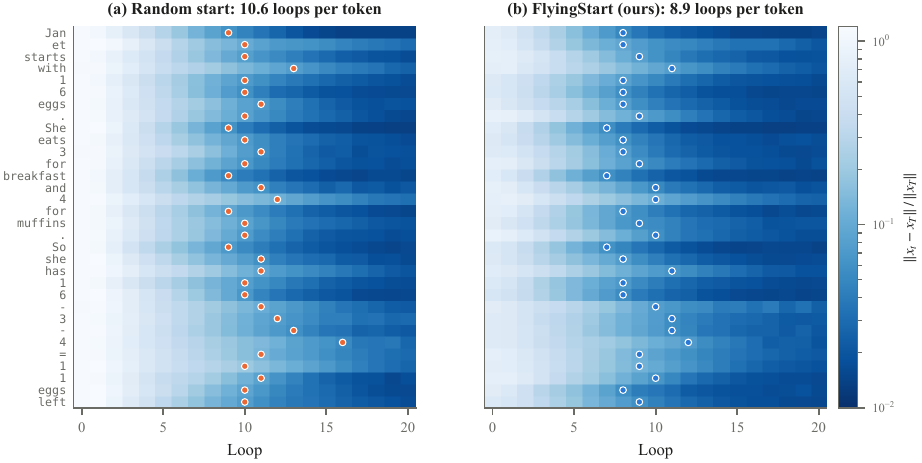}
\caption{\textbf{FlyingStart moves each token's exit earlier and keeps which tokens are hard.} Relative distance to the final state ($T = 32$) per loop for the first 32 tokens Huginn generates after one GSM8K prompt, replayed from the random start (a) and from FlyingStart (b); dots: exit loop at $\epsilon = 0.085$.}
\label{fig:tokmap}
\end{figure*}

\paragraph{Token by token.}\label{app:tokmap}
Figure~\ref{fig:tokmap} shows the distance to the final state for the first 32 tokens Huginn generates after one GSM8K prompt, replayed at full depth from each start (rerun on an RTX 3090; tokens and exit loops agree with the replay of Section~\ref{sec:analysis} to within one loop).
From the random start every token begins at the same distance (1.17); FlyingStart starts each token at its own distance (0.51--1.01) and moves most exits one or two loops earlier.
Over the 122 tokens of two prompts, 101 exit earlier, 18 at the same loop and 3 later (1.6 loops on average at $\epsilon = 0.085$), and the exit loops of the two starts correlate at 0.64: tokens that converge slowly from noise, such as the numbers in the subtraction, also converge slowly from FlyingStart.

\paragraph{Criteria fixed before each run.}
Before each experiment we wrote down a pass/fail criterion in our research log. Table~\ref{tab:prereg} lists all criteria relevant to this paper, including those that failed and one decision we later reversed.

\begin{table*}[t]
\centering\small
\begin{tabular}{p{0.54\linewidth}p{0.14\linewidth}p{0.245\linewidth}}
\toprule
Criterion (written before the run) & Outcome & Consequence \\
\midrule
Training-free solver changes (Anderson, extrapolation, over-relaxation, copy warm start): no more disagreement than the baseline at $\geq$25\% fewer layers & failed & solver-side acceleration dropped \\
LoRA jump: relative error $\leq 0.32$ after one adapted step & failed (0.65/0.54) & dropped \\
Trained (MLP) initializer: error after 4 loops $\leq$ model's after 7 & failed & linear map kept \\
Exit and stop decoupled: $\geq$15\% fewer layers or $+1.5$ points at equal layers & failed & dropped \\
Fixed depth with initializer: curve shifted by $\geq$1.5 loops; lossless fixed depth $\leq 9$ & both failed & adaptive exit kept \\
Few-step distilled student: $K\leq 6$ loops at $\geq$41.76\% on GSM8K & failed (34.1\%) & distillation dropped \\
Ablation: full initializer more accurate than each single-input variant & failed & switched to previous-state map; reverted, since it exits later (41.4 vs.\ 40.1 layers) and all main results use the full map \\
Second model: curve above the strongest baseline, pooled & passed & two-model claim, with boundary \\
Code in fit data: code gain $\geq$ half of Wikipedia gain, Wikipedia drop $\leq 0.2$ & passed (Huginn), failed (Llama) & reported as fit-data effect \\
Code-fit initializer on math: drop $\leq 0.3$ points & failed ($-0.41$, n.s.) & main results keep text+math fit \\
Format fit, zero-shot and code tasks pooled: at 10 fixed loops within 0.3 points of the baseline at 10; at $\epsilon=0.085$ within 0.3 points of it at $\leq 40$ layers; 8-shot math drop $\leq 0.3$ & failed ($-1.37$); failed ($-1.73$); passed & default fit kept; no gain on these tasks reported \\
Held-out selection, first grid (rule A, $\tau=1$): selection $\geq$10\% cheaper in $\geq$90\% of splits; median held-out loss within 0.3 points of the baseline's & first passed (19.4\%, 99.8\%), second failed (1.14 vs.\ 0.24) & symmetric grid run \\
Symmetric grid: full-sample operating point $\geq$10\% cheaper; rule A as above with the mean held-out loss; rule B inconclusive if either method falls back in $>$20\% of splits & passed (14.3\%); passed (14.4\%, 99.9\%; $+0.08$); inconclusive (29\%) & operating point and Figure~\ref{fig:curve} from the symmetric grid \\
Operating point, seeds 1 and 2: $\geq$10\% fewer layers at every seed; three-seed accuracy within 0.3 points of the baseline's & passed (14.2--14.3\%); passed ($+0.29$) & three-seed numbers reported; operating point not reselected \\
WaveFront self-speculative decoding, GSM8K test: $\geq$8\% higher throughput with full-depth verification and with the exit; accuracy $\geq -1$ point and $z > -1.96$ & passed ($\times$1.10, $-0.08$); passed ($\times$1.14, $+0.15$) & Section~\ref{sec:serving}, Table~\ref{tab:spec} \\
Same on an H100, first 330 test problems: both throughput ratios $\geq 1.08$ & passed ($\times$1.12, $\times$1.18) & both GPUs reported \\
WaveFront on Spec-Bench (out of domain): throughput ratio $\geq 1.05$ with full-depth verification; $\geq 1.05$ with the exit & failed ($\times$1.04); passed ($\times$1.24) & both reported, Table~\ref{tab:spec_extra} \\
WaveFront with speculative sampling ($T=0.7$, GSM8K): ratio $\geq 1.08$, accuracy $\geq -1$ point and $z > -1.96$ & passed ($\times$1.13, $-0.30$) & reported \\
Late entry without fine-tuning: FlyingStart on a context completed to 32 loops, 8 loops per token, $\geq +0.5$ points and $z > 1.96$ & passed ($+2.03$, $z=4.64$) & Appendix~\ref{app:late} \\
Exits on the output distribution (thresholds chosen on GSM8K-train; minimum depth added after the first pilot): any configuration $\geq$5\% fewer layers than the state-change frontier at matched accuracy, CI above 0 & not met (best $-45\%$) & state change kept as strongest baseline; Table~\ref{tab:conf} \\
BBH and IFEval, each: at $\epsilon=0.085$ layers $\leq 0.90\times$ the baseline's, accuracy $\geq -1$ point and $z > -1.96$ & passed ($\times$0.86, $-0.11$; $\times$0.75, $+0.18$) & Section~\ref{sec:text}, Table~\ref{tab:nonmath} \\
TinyLlama (fourth model), recorded before its evaluation: engine agreement $\geq 99\%$; saving at equal threshold within $0.71 \pm 0.3$ loops at all four $\epsilon$; $\geq +1$ point ($z > 1.96$) at two loops; $|\Delta| < 0.5$ and $|z| < 1.96$ at five and six loops & passed (100\%); passed (0.56--0.85); passed ($+3.49$); passed ($-0.22$, $-0.26$) & Section~\ref{sec:second}, Table~\ref{tab:tiny} \\
Calibration size: map fitted on 160 of the 640 sequences, at $\epsilon=0.085$ layers $\leq 0.90\times$ the baseline's, accuracy $\geq -1$ point and $z > -1.96$ & passed ($\times$0.86, $+0.24$) & Section~\ref{sec:method}, Figure~\ref{fig:calsize} \\
Prefill: parallel (prelude-output) start minus random start at some prefill depth in $\{8, 12, 16, 24\}$, $\geq +0.5$ points and $z > 1.96$ & passed at 8 loops ($+1.79$, $z=3.29$); n.s.\ from 12 & Section~\ref{sec:serving}, Table~\ref{tab:prefill} \\
Chat in the calibration set, CDB replay: throughput gain over the random start $\geq 2$ points above the default map's at $\geq 3$ of the four thresholds & failed (1.7--2.0; one of four) & reported in Appendix~\ref{app:details} only \\
\bottomrule
\end{tabular}
\caption{Criteria written before each run, and outcomes.}
\label{tab:prereg}
\end{table*}

\paragraph{Operating point selected on held-out problems.}\label{app:select}
Section~\ref{sec:huginn} compares each method's cheapest configuration that is not significantly below full depth, chosen on the full sample.
Here the choice is made on one half of the problems and reported on the other.
Candidates are the ten configurations of each method on the symmetric grid (state-change exit at $\epsilon \in \{0.07, 0.085, 0.1, 0.12, 0.15\}$ and fixed depths 8--12; seed 0, RTX 3090), and each of 1,000 splits draws half of every task for selection.
Rule A selects a method's cheapest candidate whose accuracy loss against full depth on the selection half is at most $\tau$ points; rule B requires the one-sided 95\% upper bound of the paired loss to be at most $\tau$; both fall back to full depth.

A first version of this analysis used the configurations available at the time, on which the baseline had no setting between $\epsilon = 0.07$ and $0.1$ and no fixed depth of 9 or 11.
It gave an 18\% saving, and its criterion that the median held-out loss of FlyingStart's selection lie within 0.3 points of the baseline's failed (1.14 against 0.24 points): rule A chose $\epsilon = 0.1$ for FlyingStart, whose full-sample loss of 0.98 points sits at the tolerance, and $\epsilon = 0.07$ for the baseline, so the medians fell on configurations of different accuracy.
We then ran the symmetric grid and, before running it, fixed the rules, $\tau = 1$, and the criteria: a saving of at least 10\% with FlyingStart cheaper in at least 90\% of splits, and a mean held-out loss within 0.3 points of the baseline's.
We changed the statistic from the median to the mean after seeing the first analysis, in which the mean difference was $+0.25$ points; Table~\ref{tab:select} reports both.
The tolerances other than $\tau = 1$ were added afterwards.

At $\tau = 1$, rule A selects $\epsilon = 0.085$ for the baseline in 59\% of splits ($\epsilon = 0.1$ in 28\%) and fixed depth 9 or $\epsilon = 0.085$ for FlyingStart in 39\% and 36\% ($\epsilon = 0.1$ in 20\%).
FlyingStart's selection is cheaper in 99.9\% of the splits and saves 14.4\% of layers on average (39.5 against 46.1), at a mean held-out loss of 0.92 points against 0.84 ($+0.08$; medians 0.94 and 0.55), so both criteria pass.
Across the four tolerances, rule A saves 13.8--16.3\% on average at mean held-out losses between 0.22 points lower and 0.08 points higher than the baseline's.
Rule B at $\tau = 1$ falls back to full depth in 29\% of splits for the baseline and 6\% for FlyingStart: the standard error of the loss on half of the problems is about 0.5 points, so a configuration is certified only when its loss on the selection half is below about 0.15 points.
Our criterion declared rule B inconclusive when either method falls back in more than 20\% of splits.
In the 70\% of splits in which it certifies a configuration of both methods, it selects 45.0 against 50.3 layers on average (10.5\%) at held-out losses of 0.45 and 0.48 points; at $\tau = 1.5$ and $2$ the saving is 14.7\% and 14.2\%.

\input{tables/select}

\paragraph{Faster-converging models, matched accuracy.}
On the Llama model, inverting the baseline frontier gives 15.1\% fewer layers at $\epsilon=0.3$ (95\% CI $[12.1, 18.3]$), 12.3\% at $\epsilon=0.2$ ($[8.4, 20.3]$) and 14.3\% at two fixed loops ($[11.3, 17.3]$). The savings are not significant at $\epsilon \leq 0.15$ or three or more fixed loops.
On OLMo the same inversion gives 13.4\% at $\epsilon=0.3$ ($[8.7, 18.0]$), 7.7\% at $\epsilon=0.2$ ($[4.1, 18.0]$) and 14.2\% at two fixed loops ($[9.1, 18.6]$); at $\epsilon \le 0.15$ and three or more fixed loops the intervals include zero. Over the points with three or more loops the curve gap is $+0.23$ ($[-0.05, +0.42]$).

\paragraph{Seeds.}\label{app:seeds}
We reran eleven Huginn configurations, the baseline frontier and our curve as they stood before the symmetric grid, with two more seeds of the random initial state on all four tasks (Table~\ref{tab:seeds}); all runs in this table are on H100.
The initializer is the same in all runs; with FlyingStart the seed still affects the random start of every prompt token during prefill, and through it the first generated token and the context of all later ones.
Paired tests use the per-problem mean over the three seeds (sign test); savings use the same task-stratified bootstrap over problems as the main text, with a problem's three seeds resampled together.
Across the three seeds, pooled accuracy has a standard deviation of 0.05--0.77 points per configuration.
\begin{itemize}
\item \textbf{Equal settings.} FlyingStart $-$ baseline at $\epsilon = 0.07$: $+0.27$ points ($z=+0.73$), with 9.76 versus 11.43 loops per token. FlyingStart at ten fixed loops $-$ baseline at twelve: $-0.35$ points ($z=-0.53$).
\item \textbf{Similar loops.} FlyingStart at $\epsilon=0.12$ $-$ baseline at $\epsilon=0.2$ (7.78 versus 7.64 loops): $+3.33$ points ($z=+7.61$). FlyingStart at $\epsilon=0.1$ $-$ baseline at $\epsilon=0.15$ (8.43 versus 8.63): $+1.74$ points ($z=+4.68$).
\item \textbf{Curve gap.} Against the baseline frontier (state-change exit at four $\epsilon$ and fixed depth 8, 10, 12), our four points lie $+2.01$ points higher at equal loops on average ($[+1.70, +2.35]$); seed 0 alone gives $+1.78$ ($[+1.39, +2.19]$).
\item \textbf{Matched accuracy.} On the same frontier, three seeds give 16.6\% fewer layers at $\epsilon=0.1$ ($[14.1, 23.5]$), 18.9\% at $\epsilon=0.12$ ($[13.0, 20.7]$) and 8.1\% at ten fixed loops ($[0.6, 14.1]$); seed 0 alone gives 16.8\%, 15.8\% and 7.5\%. The bootstrap intervals are asymmetric because the frontier is piecewise linear.
\end{itemize}

\begin{table*}[t]
\centering
\small
\begin{tabular}{lcccccc}
\toprule
Configuration & Seed 0 & Seed 1 & Seed 2 & Mean & SD & Layers \\
\midrule
Baseline, $\epsilon=0.07$ & 61.46 & 60.93 & 60.83 & 61.07 & 0.34 & 49.7 \\
Baseline, $\epsilon=0.1$ & 60.24 & 60.50 & 60.30 & 60.35 & 0.14 & 44.3 \\
Baseline, $\epsilon=0.15$ & 58.82 & 58.80 & 58.53 & 58.72 & 0.16 & 38.5 \\
Baseline, $\epsilon=0.2$ & 56.32 & 56.42 & 56.36 & 56.37 & 0.05 & 34.5 \\
Fixed, 12 loops & 61.42 & 61.07 & 61.03 & 61.17 & 0.21 & 52.0 \\
Fixed, 10 loops & 60.26 & 59.59 & 59.71 & 59.85 & 0.36 & 44.0 \\
Fixed, 8 loops & 57.88 & 56.66 & 56.46 & 57.00 & 0.77 & 36.0 \\
\midrule
FlyingStart (ours), $\epsilon=0.07$ & 61.62 & 61.17 & 61.24 & 61.34 & 0.24 & 43.1 \\
FlyingStart (ours), $\epsilon=0.1$ & 60.54 & 60.20 & 60.65 & 60.46 & 0.23 & 37.7 \\
FlyingStart (ours), $\epsilon=0.12$ & 59.65 & 60.02 & 59.43 & 59.70 & 0.30 & 35.1 \\
FlyingStart (ours), fixed 10 loops & 61.01 & 60.93 & 60.54 & 60.82 & 0.25 & 44.0 \\
\bottomrule
\end{tabular}
\caption{Huginn accuracy (\%) pooled over the four tasks for three seeds of the random initial state. Layers per generated token are averaged over seeds.}
\label{tab:seeds}
\end{table*}

\paragraph{Diffusion-forcing sampler.}\label{app:ds}
The sampler of \citet{parallelsamplers2025} iterates a wavefront of up to 128 positions jointly, runs $r'$ loops per step, appends one new position per step and freezes positions whose state changes by less than $0.03$ (relative) in a step.
We port the authors' implementation and change only the starting state of each new position: the model's random start (their default), a copy of the left neighbour (their \texttt{continuous\_compute} option), or FlyingStart applied to the new position's input and its left neighbour's current state.
Table~\ref{tab:ds} gives the results.
FlyingStart changes accuracy by $-0.91$, $+0.53$ and $+0.45$ points at $r' = 4, 2, 1$ (McNemar $p = 0.15$, $0.48$, $0.63$), so it does not recover the 4.0 points the sampler loses from $r'=4$ to $r'=1$.
Steps per token, and with them batch-one latency, are unchanged, because the sampler freezes at most one position per step.
Median compute per token falls by 8\%, 5\% and 2\%.
A map refitted on the sampler's own trajectories (GSM8K train, 54k frozen positions) changes accuracy by $-0.45$, $-0.23$ and $0.00$ points relative to the random start at $r'=4, 2, 1$ (McNemar $p = 0.50$, $0.81$, $1.0$) and saves 6\%, 3\% and 2\% of compute.
The gain is small because a position stays in the wavefront for about 60 loops before it freezes, and its left neighbour, the input to the map, has itself run only a few loops when the new position starts. The refitted map's held-out starting error is 0.84--0.93, against 0.65--0.73 in token-by-token decoding and 1.17 for the random start.
About 1\% of problems at $r'=4$, and 40\% at $r'=1$, fill the wavefront without freezing; these dominate total compute, so we report medians.

\input{tables/ds}

\paragraph{Fitting on code.}
Adding 160 sequences of Python files (codeparrot) to the fitting data (409k tokens) leaves the held-out error on Wikipedia and math unchanged (Llama: 0.52 and 0.49) and gives 0.50 on held-out code.
Its effect on the token-level evaluation is shown in Figure~\ref{fig:text} and Table~\ref{tab:text}; on the math tasks it changes accuracy by $-0.41$ ($\epsilon=0.085$, $z=-1.49$) and $+0.02$ (fixed 10 loops) points.

\end{document}

%% file: tables/spec.tex
\begin{table}[t]
\centering\small
\setlength{\tabcolsep}{3.2pt}
\begin{tabular}{lcccccc}
\toprule
Decoding & $d$ & Start & Acc. & Speedup & Gain & Accept \\
\midrule
\multicolumn{7}{l}{\emph{Verification at 32 loops (lossless)}} \\
Token by token & -- & rand. & 37.15 & 1.00$\times$ &  & -- \\
WaveFront & 4 & rand. & 36.39 & 2.74$\times$ &  & 0.926 \\
\rowcolor{fsrow}WaveFront & 4 & FS & 36.32 & 3.02$\times$ & +10\% & 0.957 \\
\midrule
\multicolumn{7}{l}{\emph{State-change exit, $\epsilon = 0.085$}} \\
Token by token & -- & rand. & 34.34 & 2.22$\times$ &  & -- \\
\rowcolor{fsrow}Token by token & -- & FS & 34.34 & 2.51$\times$ & +13\% & -- \\
WaveFront & 3 & rand. & 34.04 & 3.50$\times$ &  & 0.892 \\
\rowcolor{fsrow}WaveFront & 2 & FS & 34.19 & 3.98$\times$ & +14\% & 0.912 \\
\bottomrule
\end{tabular}
\caption{\textbf{FlyingStart (FS, ours) adds 10--14\% to WaveFront self-speculative decoding \citep{wavefront2026} at unchanged accuracy.} Huginn, GSM8K test (1,319 problems), batch size 1, RTX 3090. $d$: draft depth. Speedup over token-by-token decoding at 32 loops (5.6 tokens/s); gain over the random start (rand.) with the same decoding. Accept: fraction of drafts accepted.}
\label{tab:spec}
\end{table}

%% file: tables/prefill.tex
\begin{table}[t]
\centering\small
\setlength{\tabcolsep}{3.5pt}
\begin{tabular}{ccccc}
\toprule
& \multicolumn{2}{c}{Prefill start} & & \\
\cmidrule(lr){2-3}
$T_p$ & Random & Parallel (ours) & $z$ & Layers per problem \\
\midrule
8 & 53.29 & 55.08 & +3.29 & 31k ($-$71\%) \\
12 & 59.26 & 58.86 & $-$0.93 & 44k ($-$59\%) \\
16 & 60.10 & 60.20 & +0.29 & 57k ($-$47\%) \\
24 & 60.83 & 60.91 & +0.28 & 82k ($-$24\%) \\
\midrule
32 & 61.22 & -- & -- & 108k ($-$0.4\%) \\
\bottomrule
\end{tabular}
\caption{\textbf{Prefill depth and prefill start on Huginn.} Accuracy (\%) with $T_p$ prefill loops, pooled over the four math tasks (5,078 problems), seed 0, RTX 3090; decoding is FlyingStart at $\epsilon = 0.085$ throughout. The parallel start is the map from the prelude output alone, the only part of FlyingStart available when all prompt positions are processed at once; $z$: sign test of the parallel against the random start. Layers per problem count prefill ($4 + 4T_p$ per prompt token) and decoding; in parentheses, the change against the baseline with 32 prefill loops (108k, 61.05\%).}
\label{tab:prefill}
\end{table}

%% file: tables/huginn_tasks.tex
\begin{table*}[t]
\centering\small
\begin{tabular}{lrrrrrrr}
\toprule
Configuration & GSM8K & SVAMP & MultiArith & ASDiv & Pooled & Loops & Layers \\
\midrule
\multicolumn{8}{l}{\emph{Reference}} \\
Full depth (32 loops) & 43.37 & 59.10 & 85.83 & 66.98 & 61.52 & 32.00 & 132.0 \\
Fixed 8 & 38.59 & 56.10 & 81.67 & 63.87 & 57.88 & 8.00 & 36.0 \\
Fixed 10 & 41.62 & 58.40 & 84.67 & 65.72 & 60.26 & 10.00 & 44.0 \\
Fixed 12 & 43.29 & 59.10 & 85.83 & 66.79 & 61.42 & 12.00 & 52.0 \\
\midrule
\multicolumn{8}{l}{\emph{Strongest baseline (state-change exit, 1 KV entry)}} \\
$\epsilon$ = 0.07 & 43.21 & 60.00 & 85.33 & 66.65 & 61.46 & 11.44 & 49.8 \\
$\epsilon$ = 0.1 & 41.62 & 58.30 & 84.50 & 65.77 & 60.24 & 10.09 & 44.3 \\
$\epsilon$ = 0.15 & 40.03 & 57.70 & 81.83 & 64.43 & 58.82 & 8.63 & 38.5 \\
$\epsilon$ = 0.2 & 36.24 & 55.40 & 81.33 & 62.07 & 56.32 & 7.64 & 34.6 \\
\midrule
\multicolumn{8}{l}{\emph{FlyingStart (ours)}} \\
$\epsilon$ = 0.07 & 42.91 & 59.30 & 85.83 & 67.39 & 61.62 & 9.76 & 43.1 \\
$\epsilon$ = 0.085 & 42.30 & 59.20 & 86.50 & 66.93 & 61.32 & 9.03 & 40.1 \\
$\epsilon$ = 0.1 & 41.09 & 58.80 & 85.50 & 66.28 & 60.54 & 8.43 & 37.7 \\
$\epsilon$ = 0.12 & 39.50 & 57.50 & 85.00 & 65.91 & 59.65 & 7.78 & 35.1 \\
Fixed 8 & 39.35 & 58.60 & 84.83 & 64.75 & 59.31 & 8.00 & 36.0 \\
Fixed 9 & 40.71 & 58.90 & 84.50 & 66.56 & 60.46 & 9.00 & 40.0 \\
Fixed 10 & 42.53 & 58.90 & 85.17 & 66.56 & 61.01 & 10.00 & 44.0 \\
\bottomrule
\end{tabular}
\caption{Huginn-0125, accuracy (\%) per task and pooled (5,078 problems). Loops: mean per generated token, weighted by tokens. Layers $= 4 + 4\times$loops.}
\label{tab:huginn}
\end{table*}

%% file: tables/retro_tasks.tex
\begin{table*}[t]
\centering\small
\begin{tabular}{lrrrrrrr}
\toprule
Configuration & GSM8K & SVAMP & MultiArith & ASDiv & Pooled & Loops & Layers \\
\midrule
\multicolumn{8}{l}{\emph{Reference}} \\
Full depth (32 loops) & 55.42 & 67.60 & 93.00 & 83.14 & 74.04 & 32.00 & 200.0 \\
Fixed 2 & 42.08 & 59.30 & 90.50 & 77.44 & 66.23 & 2.00 & 20.0 \\
Fixed 3 & 50.95 & 66.10 & 93.50 & 80.96 & 71.72 & 3.00 & 26.0 \\
Fixed 4 & 54.44 & 67.00 & 92.33 & 82.21 & 73.20 & 4.00 & 32.0 \\
Fixed 5 & 54.51 & 66.90 & 92.00 & 82.49 & 73.28 & 5.00 & 38.0 \\
Fixed 6 & 55.57 & 67.40 & 92.83 & 82.91 & 73.93 & 6.00 & 44.0 \\
\midrule
\multicolumn{8}{l}{\emph{Strongest baseline}} \\
$\epsilon$ = 0.1 & 53.98 & 66.80 & 92.00 & 82.12 & 72.96 & 4.18 & 33.1 \\
$\epsilon$ = 0.15 & 53.83 & 66.50 & 92.17 & 81.33 & 72.55 & 3.57 & 29.4 \\
$\epsilon$ = 0.2 & 51.48 & 66.70 & 93.67 & 81.84 & 72.37 & 3.17 & 27.0 \\
$\epsilon$ = 0.3 & 49.96 & 65.70 & 93.17 & 81.15 & 71.43 & 2.91 & 25.5 \\
\midrule
\multicolumn{8}{l}{\emph{FlyingStart (ours)}} \\
$\epsilon$ = 0.1 & 53.68 & 67.00 & 92.67 & 82.17 & 73.02 & 3.57 & 29.4 \\
$\epsilon$ = 0.15 & 53.15 & 66.60 & 92.17 & 81.38 & 72.41 & 3.01 & 26.1 \\
$\epsilon$ = 0.2 & 51.55 & 67.10 & 91.17 & 81.19 & 71.90 & 2.51 & 23.0 \\
$\epsilon$ = 0.3 & 48.22 & 65.20 & 91.67 & 79.62 & 70.05 & 2.07 & 20.4 \\
Fixed 2 & 46.47 & 65.30 & 91.17 & 79.20 & 69.38 & 2.00 & 20.0 \\
Fixed 3 & 53.07 & 67.90 & 91.67 & 80.59 & 72.25 & 3.00 & 26.0 \\
Fixed 4 & 54.51 & 66.80 & 92.33 & 82.17 & 73.16 & 4.00 & 32.0 \\
Fixed 5 & 55.04 & 66.80 & 92.00 & 82.35 & 73.34 & 5.00 & 38.0 \\
Fixed 6 & 55.50 & 67.20 & 92.33 & 82.68 & 73.71 & 6.00 & 44.0 \\
\bottomrule
\end{tabular}
\caption{Retrofitted Llama-3.2-1B, accuracy (\%) per task and pooled. Layers $= 8 + 6\times$loops.}
\label{tab:retro}
\end{table*}

%% file: tables/tiny_tasks.tex
\begin{table*}[t]
\centering\small
\begin{tabular}{lrrrrrrr}
\toprule
Configuration & GSM8K & SVAMP & MultiArith & ASDiv & Pooled & Loops & Layers \\
\midrule
\multicolumn{8}{l}{\emph{Reference}} \\
Full depth (32 loops) & 45.64 & 64.10 & 87.50 & 78.09 & 68.02 & 32.00 & 264.0 \\
Fixed 2 & 33.66 & 56.20 & 79.50 & 72.07 & 59.85 & 2.00 & 24.0 \\
Fixed 3 & 41.62 & 62.00 & 89.33 & 77.03 & 66.33 & 3.00 & 32.0 \\
Fixed 4 & 44.28 & 63.10 & 89.17 & 77.54 & 67.43 & 4.00 & 40.0 \\
Fixed 5 & 44.28 & 64.00 & 88.50 & 77.81 & 67.64 & 5.00 & 48.0 \\
Fixed 6 & 45.03 & 64.00 & 88.17 & 78.14 & 67.94 & 6.00 & 56.0 \\
\midrule
\multicolumn{8}{l}{\emph{Strongest baseline}} \\
$\epsilon$ = 0.1 & 44.66 & 63.20 & 89.50 & 77.72 & 67.66 & 4.36 & 42.9 \\
$\epsilon$ = 0.15 & 44.12 & 62.70 & 89.67 & 77.81 & 67.49 & 3.83 & 38.6 \\
$\epsilon$ = 0.2 & 44.35 & 61.50 & 89.17 & 76.70 & 66.78 & 3.37 & 35.0 \\
$\epsilon$ = 0.3 & 42.08 & 62.00 & 88.83 & 76.52 & 66.17 & 3.03 & 32.2 \\
\midrule
\multicolumn{8}{l}{\emph{FlyingStart (ours)}} \\
$\epsilon$ = 0.1 & 44.43 & 64.00 & 89.00 & 77.67 & 67.68 & 3.80 & 38.4 \\
$\epsilon$ = 0.15 & 45.64 & 63.40 & 87.50 & 77.30 & 67.55 & 3.24 & 33.9 \\
$\epsilon$ = 0.2 & 44.28 & 63.20 & 89.33 & 76.89 & 67.19 & 2.78 & 30.2 \\
$\epsilon$ = 0.3 & 40.56 & 61.00 & 87.17 & 74.80 & 64.65 & 2.18 & 25.4 \\
Fixed 2 & 37.98 & 60.00 & 86.00 & 74.06 & 63.33 & 2.00 & 24.0 \\
Fixed 3 & 44.12 & 63.80 & 88.67 & 76.56 & 67.05 & 3.00 & 32.0 \\
Fixed 4 & 44.81 & 62.60 & 88.67 & 77.30 & 67.31 & 4.00 & 40.0 \\
Fixed 5 & 44.50 & 63.60 & 87.17 & 77.72 & 67.43 & 5.00 & 48.0 \\
Fixed 6 & 44.88 & 63.40 & 87.83 & 78.00 & 67.68 & 6.00 & 56.0 \\
\bottomrule
\end{tabular}
\caption{Retrofitted TinyLlama-1.1B, accuracy (\%) per task and pooled (RTX A4500). Layers $= 8 + 8\times$loops. The saving (0.71 loops) and the shape of the curve were predicted before these runs.}
\label{tab:tiny}
\end{table*}

%% file: tables/text.tex
\begin{table*}[t]
\centering\small
\begin{tabular}{llrrrrrr}
\toprule
& & \multicolumn{3}{c}{Wikipedia} & \multicolumn{3}{c}{Code (MBPP)} \\
\cmidrule(lr){3-5}\cmidrule(lr){6-8}
Model & Configuration & Layers & Flip (\%) & NLL & Layers & Flip (\%) & NLL \\
\midrule
Huginn & Full depth & 132.0 & 0.00 & 2.1352 & 132.0 & 0.00 & 1.0967 \\
 & Baseline $\epsilon$=0.05 & 51.2 & 2.87 & 2.1349 & 53.0 & 1.41 & 1.0970 \\
 & Baseline $\epsilon$=0.07 & 46.4 & 3.60 & 2.1364 & 47.9 & 1.87 & 1.0974 \\
 & Baseline $\epsilon$=0.1 & 41.6 & 4.67 & 2.1385 & 42.8 & 2.54 & 1.0984 \\
 & Baseline $\epsilon$=0.15 & 36.3 & 6.34 & 2.1464 & 37.2 & 3.71 & 1.1048 \\
 & Baseline $\epsilon$=0.2 & 32.7 & 8.27 & 2.1561 & 33.3 & 4.72 & 1.1137 \\
 & FlyingStart (ours) $\epsilon$=0.05 & 46.5 & 2.89 & 2.1346 & 50.3 & 1.49 & 1.0969 \\
 & FlyingStart (ours) $\epsilon$=0.07 & 41.7 & 3.47 & 2.1363 & 45.2 & 1.80 & 1.0969 \\
 & FlyingStart (ours) $\epsilon$=0.1 & 36.9 & 4.59 & 2.1389 & 40.1 & 2.55 & 1.0993 \\
 & FlyingStart (ours) $\epsilon$=0.15 & 31.8 & 6.09 & 2.1456 & 34.6 & 3.64 & 1.1047 \\
 & FlyingStart (ours) $\epsilon$=0.2 & 28.2 & 7.96 & 2.1557 & 30.9 & 4.61 & 1.1124 \\
 & FlyingStart (ours) $\epsilon$=0.05, +code fit & 46.6 & 2.98 & 2.1344 & 48.8 & 1.48 & 1.0966 \\
 & FlyingStart (ours) $\epsilon$=0.1, +code fit & 37.1 & 4.61 & 2.1396 & 38.7 & 2.51 & 1.0991 \\
 & FlyingStart (ours) $\epsilon$=0.2, +code fit & 28.3 & 8.00 & 2.1555 & 29.6 & 4.61 & 1.1125 \\
\midrule
Llama & Full depth & 200.0 & 0.00 & 3.0248 & 200.0 & 0.00 & 2.0512 \\
 & Baseline $\epsilon$=0.1 & 31.0 & 3.81 & 3.0258 & 33.0 & 1.95 & 2.0511 \\
 & Baseline $\epsilon$=0.15 & 27.6 & 5.50 & 3.0292 & 29.2 & 2.82 & 2.0528 \\
 & Baseline $\epsilon$=0.2 & 26.3 & 6.31 & 3.0311 & 27.2 & 3.47 & 2.0560 \\
 & Baseline $\epsilon$=0.3 & 23.7 & 8.98 & 3.0427 & 24.5 & 5.17 & 2.0645 \\
 & FlyingStart (ours) $\epsilon$=0.1 & 28.0 & 3.92 & 3.0249 & 32.2 & 1.99 & 2.0513 \\
 & FlyingStart (ours) $\epsilon$=0.15 & 24.7 & 5.53 & 3.0284 & 28.8 & 2.70 & 2.0522 \\
 & FlyingStart (ours) $\epsilon$=0.2 & 22.0 & 7.54 & 3.0342 & 26.0 & 3.90 & 2.0570 \\
 & FlyingStart (ours) $\epsilon$=0.3 & 20.2 & 9.48 & 3.0429 & 21.4 & 7.18 & 2.0863 \\
 & FlyingStart (ours) $\epsilon$=0.1, +code fit & 28.0 & 3.84 & 3.0250 & 30.4 & 1.88 & 2.0517 \\
 & FlyingStart (ours) $\epsilon$=0.2, +code fit & 22.0 & 7.60 & 3.0340 & 23.9 & 4.35 & 2.0634 \\
\bottomrule
\end{tabular}
\caption{Token-level evaluation under teacher forcing (128-token context, 256 decoded tokens per sequence). Flip: fraction of tokens whose greedy prediction differs from full depth. NLL: negative log-likelihood of the reference token. ``+code fit'': initializer fitted with 160 additional code sequences.}
\label{tab:text}
\end{table*}

%% file: tables/throughput.tex
\begin{table}[h]
\centering\small
\setlength{\tabcolsep}{4pt}
\begin{tabular}{lrrrr}
\toprule
& \multicolumn{2}{c}{Batch 32} & \multicolumn{2}{c}{Batch 128} \\
\cmidrule(lr){2-3}\cmidrule(lr){4-5}
Configuration & Sync & Async & Sync & Async \\
\midrule
Full depth & 206 & -- & 420 & -- \\
Baseline $\epsilon$=0.07 & 259 & 296 & 411 & 732 \\
Baseline $\epsilon$=0.1 & 312 & 334 & 483 & 808 \\
FlyingStart (ours) $\epsilon$=0.085 & 305 & 358 & 508 & 871 \\
FlyingStart (ours) $\epsilon$=0.1 & 327 & 388 & 542 & 921 \\
Fixed 12 & 499 & -- & 997 & -- \\
FlyingStart (ours) + fixed 10 & 578 & 561 & 1155 & 1057 \\
\bottomrule
\end{tabular}
\caption{Decoding throughput of Huginn-0125 (tokens/s, one H100, about 850-token prompts, 64 generated tokens per row). Sync: all rows of a batch advance loop by loop together. Async: our implementation of depth-asynchronous batching \citep{cdb2026}.}
\label{tab:throughput}
\end{table}

%% file: tables/m3.tex
\begin{table*}[t]
\centering\small
\begin{tabular}{lccccrr}
\toprule
Configuration & GSM8K (0-shot) & MATH-500 & MBPP & HumanEval & Pooled & Layers \\
\midrule
Full depth & 33.51 & 19.80 & 29.26 & 19.51 & 28.97 & 132.0 \\
Fixed 12 & 34.72 & 18.60 & 28.46 & 20.12 & 29.25 & 52.0 \\
Fixed 10 & 33.66 & 19.80 & 28.66 & 23.17 & 29.17 & 44.0 \\
Baseline, $\epsilon=0.07$ & 32.98 & 20.80 & 28.86 & 21.95 & 28.97 & 49.1 \\
Baseline, $\epsilon=0.1$ & 32.83 & 18.60 & 28.66 & 23.17 & 28.49 & 44.4 \\
\midrule
FlyingStart (ours), fixed 10 & 32.83 & 17.60 & 27.66 & 21.34 & 27.96 & 44.0 \\
FlyingStart (ours), $\epsilon=0.085$ & 32.68 & 19.00 & 29.26 & 20.73 & 28.44 & 39.2 \\
FlyingStart (ours), $\epsilon=0.1$ & 32.60 & 18.00 & 27.86 & 21.34 & 27.96 & 37.0 \\
\midrule
+ format fit, fixed 10 & 32.15 & 16.80 & 29.46 & 21.34 & 27.80 & 44.0 \\
+ format fit, $\epsilon=0.085$ & 31.08 & 17.60 & 29.46 & 21.95 & 27.44 & 37.8 \\
+ format fit, $\epsilon=0.1$ & 32.30 & 18.60 & 28.46 & 23.17 & 28.16 & 35.7 \\
\bottomrule
\end{tabular}
\caption{Huginn-0125 on zero-shot GSM8K (CoT, 512 new tokens), MATH-500 (2,048), MBPP and HumanEval (pass@1), seed 0. Pool: 2,482 problems. Layers: decoding layers per generated token, pooled. ``+ format fit'': map refitted with 500 added sequences in the evaluation formats, chosen after seeing the rows above (Section~\ref{sec:text}).}
\label{tab:m3}
\end{table*}

%% file: tables/nonmath.tex
\begin{table}[t]
\centering\footnotesize
\setlength{\tabcolsep}{1.0pt}
\begin{tabular}{lcccc}
\toprule
& \multicolumn{2}{c}{BBH (3-shot CoT)} & \multicolumn{2}{c}{IFEval (0-shot)} \\
\cmidrule(lr){2-3}\cmidrule(lr){4-5}
Configuration & Acc. & Layers & Acc. & Layers \\
\midrule
Full depth (32 loops) & 32.41 & 132.0 & 19.78 & 132.0 \\
Fixed 12 & 32.10 & 52.0 & 20.52 & 52.0 \\
Fixed 10 & 31.85 & 44.0 & 19.04 & 44.0 \\
Baseline, $\epsilon=0.1$ & 31.85 & 45.5 & 21.07 & 47.5 \\
Baseline, $\epsilon=0.085$ & 32.05 & 47.9 & 19.22 & 49.7 \\
\midrule
\rowcolor{fsrow}FlyingStart (ours), fixed 10 & 32.05 & 44.0 & 19.59 & 44.0 \\
\rowcolor{fsrow}FlyingStart (ours), $\epsilon=0.1$ & 31.82 & 38.9 & 18.30 & 35.1 \\
\rowcolor{fsrow}FlyingStart (ours), $\epsilon=0.085$ & 31.95 & 41.2 & 19.41 & 37.1 \\
\bottomrule
\end{tabular}
\caption{\textbf{On BBH and IFEval, FlyingStart at $\epsilon=0.085$ uses 14\% and 25\% fewer layers than the baseline at the same threshold, with accuracy within noise.} Huginn-0125, seed 0, default map (fitted on Wikipedia and 8-shot GSM8K; neither benchmark is in the fit). BBH: 27 tasks, 6,511 problems, exact match, at most 384 new tokens (an output cut off before its answer counts as wrong), RTX A4500. IFEval: 541 prompts, prompt-level strict accuracy, 1,280 new tokens, RTX 3090. Baseline: random start, same state-change exit and single KV entry. Layers: decoding layers per generated token.}
\label{tab:nonmath}
\end{table}

%% file: tables/spec_extra.tex
\begin{table*}[t]
\centering\small
\setlength{\tabcolsep}{4pt}
\begin{tabular}{lcccccc}
\toprule
& \multicolumn{3}{c}{$d=4$, 32-loop verification} & \multicolumn{3}{c}{exit at $\epsilon=0.085$} \\
\cmidrule(lr){2-4}\cmidrule(lr){5-7}
Spec-Bench category & Random & FS & Ratio & Random ($d=3$) & FS ($d=2$) & Ratio \\
\midrule
All 480 & 3.95$\times$ & 4.10$\times$ & 1.04 & 5.06$\times$ & 6.26$\times$ & 1.24 \\
\midrule
Translation & 4.28$\times$ & 4.33$\times$ & 1.01 & 5.37$\times$ & 6.76$\times$ & 1.26 \\
Summarization & 3.82$\times$ & 3.95$\times$ & 1.03 & 4.89$\times$ & 5.85$\times$ & 1.20 \\
QA & 3.89$\times$ & 4.09$\times$ & 1.05 & 5.13$\times$ & 6.29$\times$ & 1.22 \\
Math reasoning & 3.99$\times$ & 4.15$\times$ & 1.04 & 4.95$\times$ & 6.39$\times$ & 1.29 \\
RAG & 3.85$\times$ & 4.03$\times$ & 1.05 & 4.93$\times$ & 6.12$\times$ & 1.24 \\
MT-Bench & 3.83$\times$ & 4.00$\times$ & 1.04 & 5.03$\times$ & 6.07$\times$ & 1.21 \\
\bottomrule
\end{tabular}
\caption{FlyingStart (FS, ours) added to WaveFront on Spec-Bench (question ids 81--560, first turn, raw prompts, 512 new tokens), batch size 1, RTX 6000 Ada. Speedup over token-by-token decoding at 32 loops; Ratio is FS over random start. Speed only: Spec-Bench has no reference answers. Sampling at temperature 0.7 on GSM8K test (1,319 problems, $d=4$, verification at 32 loops): 2.54$\times$ (random start, acceptance 0.904, accuracy 28.20\%) and 2.86$\times$ (FlyingStart, 0.942, 27.90\%), a ratio of 1.13 ($z = -0.24$ for accuracy).}
\label{tab:spec_extra}
\end{table*}

%% file: tables/late.tex
\begin{table*}[t]
\centering\small
\begin{tabular}{ccccccccc}
\toprule
& \multicolumn{4}{c}{Context at depth $K$} & \multicolumn{3}{c}{Context completed to 32 loops} \\
\cmidrule(lr){2-5}\cmidrule(lr){6-8}
$K$ & Layers & (a) per-loop & (b) last loop & (b) + FlyingStart (ours) & Layers & (c) random & (c) + FlyingStart (ours) \\
\midrule
8 & 36 & 34.86 & 57.37 & 59.18 (+3.8) & 132 & 58.21 & 60.24 (+4.6) \\
12 & 52 & 53.60 & 61.44 & 61.64 (+0.6) & 132 & 61.56 & 61.26 (-1.0) \\
16 & 68 & 58.41 & 61.32 & 61.28 (-0.2) & 132 & 61.48 & 61.28 (-0.8) \\
\bottomrule
\end{tabular}
\caption{Late entry without fine-tuning on Huginn: every token emits after $K$ loops; accuracy (\%) pooled over the four math tasks (5,078 problems), seed 0, RTX 3090; full depth 61.48\%. (a)~per-loop KV truncated at $K$ (a new token's loop $t$ reads loop $t$ of earlier tokens), the early-exit baseline of \citet{lateentry2026}; (b)~one KV entry per token, its state after $K$ loops (our baseline); (c)~the token keeps iterating to 32 loops after it emits and its KV is overwritten, so later tokens read a full-depth context, the late-entry setting of \citet{lateentry2026} without its fine-tuning or pipeline lag. Layers: per generated token, including the loops after emission. In parentheses: sign-test $z$ of FlyingStart against the random start on the same context.}
\label{tab:late}
\end{table*}

%% file: tables/conf.tex
\begin{table*}[t]
\centering\small
\begin{tabular}{lccccccc}
\toprule
& & & & & & \multicolumn{2}{c}{Saving at matched accuracy (\%)} \\
\cmidrule(lr){7-8}
Exit signal & Threshold & Min.\ loops & Acc.\ (\%) & Loops & Layers & layers & loops only \\
\midrule
max.\ probability & $\ge$\,0.8 & 1 & 10.77 & 10.5 & 65.1 & below frontier & below frontier \\
max.\ probability & $\ge$\,0.8 & 4 & 45.98 & 10.5 & 58.9 & below frontier & below frontier \\
max.\ probability & $\ge$\,0.8 & 8 & 60.40 & 13.6 & 69.8 & $-56$ [$-67$, $-43$] & $-34$ [$-44$, $-22$] \\
entropy & $\le$\,1.0 & 1 & 8.37 & 9.5 & 58.9 & below frontier & below frontier \\
entropy & $\le$\,1.0 & 4 & 37.28 & 8.9 & 49.5 & below frontier & below frontier \\
entropy & $\le$\,1.0 & 8 & 59.14 & 11.9 & 59.5 & $-45$ [$-58$, $-35$] & $-29$ [$-41$, $-19$] \\
top-2 margin & $\ge$\,0.6 & 4 & 42.48 & 9.3 & 51.7 & below frontier & below frontier \\
top-2 margin & $\ge$\,0.8 & 4 & 48.96 & 12.1 & 68.8 & below frontier & below frontier \\
top-2 margin & $\ge$\,0.6 & 8 & 60.02 & 12.8 & 64.9 & $-47$ [$-60$, $-41$] & $-28$ [$-40$, $-22$] \\
\bottomrule
\end{tabular}
\caption{Exits on the output distribution (maximum probability, entropy in nats, gap between the two largest probabilities) on Huginn from the random start with one KV entry per token; four 8-shot math tasks pooled (5,078 problems), seed 0, RTX 3090. Each check runs the two coda layers, so Layers $= 2 + 4\cdot\text{loops} + 2\cdot\text{checks}$. Saving: one minus the configuration's cost over the cost at which the state-change-and-fixed-depth frontier (state change $\varepsilon \in \{0.07, \ldots, 0.15\}$, fixed 8--12 loops) reaches the same accuracy; negative means more compute; 95\% intervals from a task-stratified bootstrap over problems. ``Below frontier'': lower accuracy than any frontier point (fixed 8 loops, 57.37\%). Reference: state change $\varepsilon = 0.085$ reaches 61.05\% at 10.7 loops (46.8 layers).}
\label{tab:conf}
\end{table*}

%% file: tables/select.tex
\begin{table*}[t]
\centering\small
\begin{tabular}{lcccccccccc}
\toprule
 & & Fallback (\%) & Both & \multicolumn{2}{c}{Layers / token} & Layers saved (\%) & \multicolumn{2}{c}{Loss, mean} & \multicolumn{2}{c}{Loss, median} \\
Rule & $\tau$ & Base.\ / FS & select (\%) & Base. & FS & mean [2.5, 97.5 pct.] & Base. & FS & Base. & FS \\
\midrule
A & 0.5 & 7 / 2 & 93 & 48.4 & 41.7 & 13.8 [-2.7, 23.0] & 0.55 & 0.56 & 0.55 & 0.43 \\
 & 1$^\dagger$ & 0 / 0 & 100 & 46.1 & 39.5 & 14.4 [9.1, 22.8] & 0.84 & 0.92 & 0.55 & 0.94 \\
 & 1.5 & 0 / 0 & 100 & 44.4 & 37.5 & 15.4 [9.1, 20.9] & 1.41 & 1.44 & 1.42 & 1.42 \\
 & 2 & 0 / 0 & 100 & 42.9 & 35.8 & 16.3 [8.9, 20.9] & 1.92 & 1.70 & 1.97 & 1.73 \\
\midrule
B & 0.5 & 83 / 58 & 13 & 51.4 & 48.4 & 5.7 [-11.2, 23.0] & 0.66 & 0.55 & 0.51 & 0.35 \\
 & 1$^\dagger$ & 29 / 6 & 70 & 50.3 & 45.0 & 10.5 [-11.2, 23.0] & 0.48 & 0.45 & 0.31 & 0.51 \\
 & 1.5 & 2 / 0 & 98 & 48.0 & 40.9 & 14.7 [0.0, 23.0] & 0.58 & 0.55 & 0.55 & 0.39 \\
 & 2 & 0 / 0 & 100 & 45.9 & 39.4 & 14.2 [9.1, 20.8] & 0.90 & 0.95 & 0.63 & 0.98 \\
\bottomrule
\end{tabular}
\caption{Operating point selected on one half of the problems and reported on the other half (Huginn, 5,078 problems, 1,000 splits). Each rule selects the cheapest configuration of a method whose accuracy loss against full depth on the selection half is at most $\tau$ points, by point estimate (A) or by one-sided 95\% upper bound (B), and falls back to full depth when none qualifies. Base.: strongest baseline; FS: FlyingStart (ours). Fallback: share of splits in which no configuration of a method qualifies. Layers, savings and held-out losses (points against full depth) are taken over the splits in which both methods select a configuration. $^\dagger$: tolerance fixed before the analysis.}
\label{tab:select}
\end{table*}

%% file: tables/ds.tex
\begin{table}[h]
\centering\small
\setlength{\tabcolsep}{3.5pt}
\begin{tabular}{clrrrr}
\toprule
$r'$ & Start & Acc. & Steps/tok & Loops/tok & Tok/s \\
\midrule
4 & Random & 43.59 & 1.105 & 42.5 & 17.8 \\
4 & Copy & 43.06 & 1.106 & 42.3 & 18.2 \\
4 & FlyingStart (ours) & 42.68 & 1.095 & 39.0 & 18.2 \\
\midrule
2 & Random & 42.23 & 1.386 & 65.8 & 20.1 \\
2 & FlyingStart (ours) & 42.76 & 1.354 & 62.3 & 19.9 \\
\midrule
1 & Random & 39.58 & 2.067 & 119.1 & 18.8 \\
1 & FlyingStart (ours) & 40.03 & 2.049 & 116.7 & 18.2 \\
\bottomrule
\end{tabular}
\caption{Huginn-0125 with the diffusion-forcing sampler of \citet{parallelsamplers2025} on GSM8K (1,319 problems, 8-shot CoT, greedy, batch 1, one A6000). $r'$: loops per sampler step. Loops/tok: wavefront positions $\times$ loops per generated token, median over problems. Tok/s: median over problems.}
\label{tab:ds}
\end{table}